\documentclass[runningheads]{llncs}

\usepackage{eccv}

\usepackage{eccvabbrv}

\usepackage{graphicx}
\usepackage{booktabs}

\usepackage[accsupp]{axessibility}  

\usepackage{hyperref}

\usepackage{orcidlink}

\usepackage[utf8]{inputenc} 
\usepackage[T1]{fontenc}    
\usepackage{hyperref}       
\usepackage{url}            
\usepackage{booktabs}       
\usepackage{amsfonts}       
\usepackage{nicefrac}       
\usepackage{microtype}      
\usepackage{xcolor}         
\usepackage{amsmath}
\usepackage{multirow}
\usepackage{graphicx}
\usepackage{subcaption}   
\usepackage{placeins}   
\usepackage{caption}
\usepackage[table]{xcolor}
\usepackage[export]{adjustbox}
\usepackage[most]{tcolorbox}

\newtcolorbox{finding}[1]{
  enhanced, breakable,
  colback=gray!8, colframe=gray!55, boxrule=0.4pt, arc=2pt,
  left=8pt, right=8pt, top=6pt, bottom=6pt,
  fonttitle=\bfseries\sffamily, coltitle=black,
  title={Finding #1}
}

\begin{document}

\title{Sparse Attention for Dense Open-Vocabulary Prediction in CLIP}

\titlerunning{Sparse Attention in CLIP}

\author{
Fatimah Zohra\inst{1} \quad
Chen Zhao\inst{1,2} \quad
Shuming Liu\inst{1} \quad
Bernard Ghanem\inst{1} \\
}

\authorrunning{F.~Zohra et al.}

\institute{$^{1}$ King Abdullah University of Science and Technology (KAUST) \\
$^{2}$Harbin Institute of Technology, Shenzhen }

\maketitle

\begin{abstract}
Contrastive Language–Image Pre-training (CLIP) relies on softmax-based self-attention, a strictly positive distribution that assigns probability mass to every pair of tokens—even semantically irrelevant ones. While these dense softmax weights are effective for gathering broad context during pre-training, they spread attention across many low-salience tokens, producing noise that obscures the fine-grained, spatially localized cues required for dense, open-vocabulary prediction. We study an inference-time substitution of the row-wise softmax in the final visual self-attention layers with the $\alpha$-entmax transform, applied across both the standard query--key attention and self-correlation variants. Because entmax applies a data-dependent threshold that maps low scores exactly to zero, it acts as an implicit denoiser, zeroing contextually irrelevant dependencies while redistributing mass onto the most relevant tokens. We evaluate on open-vocabulary tasks---dense semantic segmentation (Pascal VOC, Pascal Context, ADE20K) and fine-grained retrieval (FG-OVD)---and find the gain from attention sparsification is proportional to how much the baseline attention spreads off the target class. 
\end{abstract}

\begin{figure}[htbp] 
  \centering 

  \begin{minipage}{0.39\linewidth}
    \adjustbox{trim=0 0 0 0, clip, width=\linewidth} {%
    \includegraphics[]{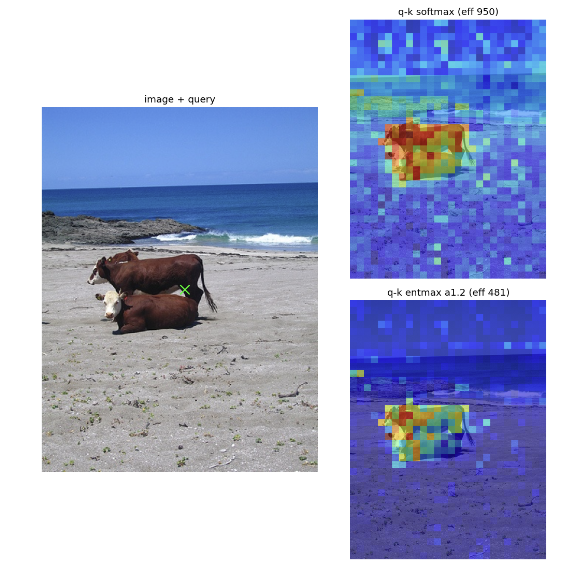}
    }
  \end{minipage}
  \hspace{0.03\linewidth} 
    \begin{minipage}{0.5\linewidth}
    \adjustbox{trim=0 {0.1\height} 0 0, clip, width=0.95\linewidth}{%
    \includegraphics{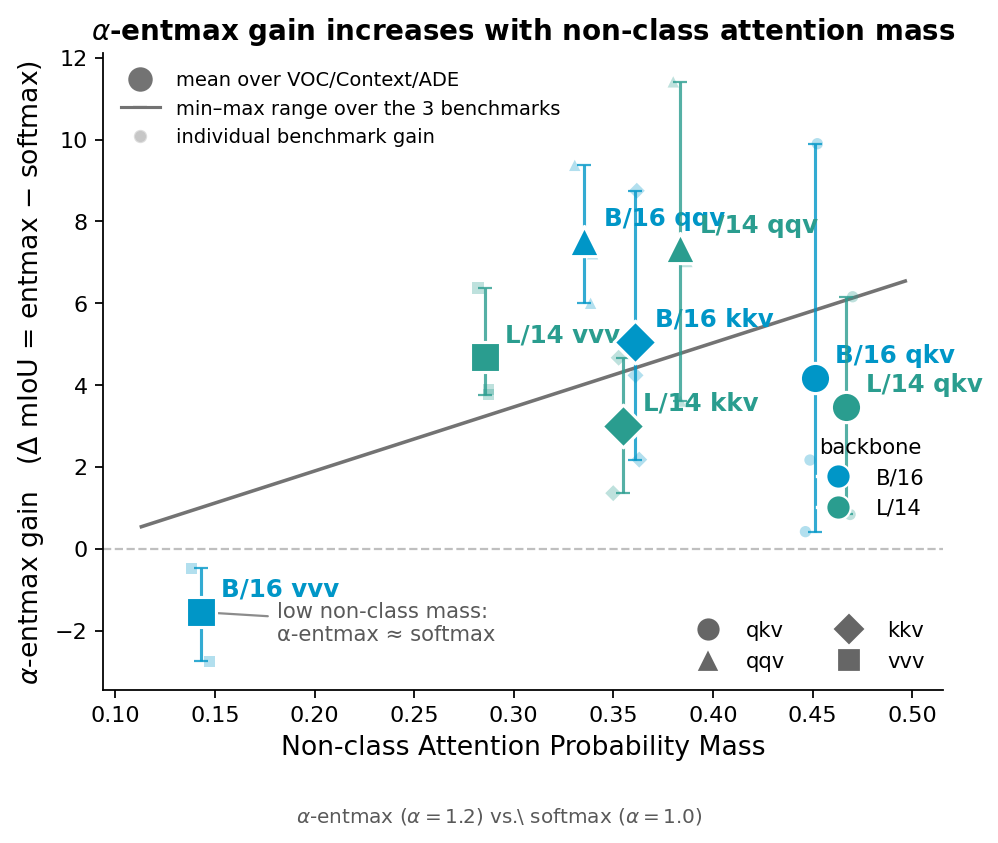}%
    }
    \end{minipage}

\caption{\textbf{Attention Mass in Softmax vs.\ Entmax Distributions.} \textbf{Left:} attention distribution for the query patch (green $\times$) under \texttt{q-k} attention. Entmax denoises the distribution, concentrating mass on the most relevant tokens --the horse's body vs. the rider-- while suppressing irrelevant background. 
\textbf{Right:}
\textit{Sparsification helps in proportion to how much the baseline attention spreads off the target class.} Where the mass is already very concentrated, entmax does not benefit. The $x$-axis shows the fraction of a foreground patch's attention that lands on non-class (irrelevant) patches, measured against the downsampled Pascal VOC ground truth and the $y$-axis is the $\alpha$-entmax ($\alpha{=}1.2$) minus softmax ($\alpha{=}1.0$) segmentation gain. $\alpha$-entmax benefits when attention is spread across non-class tokens. 
}
\label{fig:teaser}
\end{figure}

\section{Introduction}\label{sec:intro}
Transformer-based vision–language models such as CLIP \cite{radford2021learning} and its many large-scale follow-ups, e.g., EVA-CLIP \cite{eva}, SigLIP \cite{zhai2023siglip}, and ALIGN \cite{align}, all use separate image and text encoders that are trained contrastively on hundreds of millions to billions of web-scraped image–text pairs, learning a shared embedding space for both modalities. This paradigm yields task‑agnostic visual features that are aligned with language features, making CLIP the \emph{de facto} front‑end for zero‑shot and open‑vocabulary tasks.

However, CLIP's global contrastive objective aligns a global representation of an image with a short caption, optimizing the visual encoder for image-level semantics rather than the spatially localized features that dense prediction tasks require. As a result, when CLIP's patch tokens are used zero-shot for pixel- or region-level tasks, their features localize poorly: patch-level analyses show that high activations frequently land on irrelevant background regions rather than the object of interest~\cite{ranasinghe2023perceptual}, and the attention maps of the final layers grow diffuse~\cite{clip-surgery,gem}. The same image-level bias effects ROI-based fine-grained region recognition, where separating objects that differ by only a single attribute---color, material, or pattern---demands exactly the localized cues the global objective does not tease apart~\cite{yuksekgonul2023bags,ma2023crepe}. 

A growing body of work improves CLIP's dense predictions \emph{without retraining}, by modifying the final attention block exclusively. The primary observations are that dropping the query--key mixing and keeping only the value path~\cite{mask-clip}, substituting self-correlation (query--query, key--key, value--value) attention~\cite{clip-surgery,sclip,gem}, or reshaping the attention neighbourhood and simplifying the last block~\cite{naclip,lan2024clearclip} can help surface localized features. These methods establish that much of CLIP's dense-prediction noise originates in the last layers' attention---yet they address \emph{what} is correlated, but do not take into account the effect of the softmax normalizer and it's role in spreading mass globally across all tokens. 

This density is especially problematic for CLIP due to its contrastive objective ranking image--text pairs, which effectively concentrates representational variance on the directions that best separate those pairs---typically coarse object- and scene-level semantics. Fine-grained, spatially localized features are still encoded, but they lie along low-variance directions that contribute little to the image--text similarity driving training, leaving them obscured relative to the dominant coarse signal. We \textit{hypothesize} that the softmax normalizer compounds this imbalance at inference. Although the individual heads of Multi-Head Attention (MHA) tend to specialize in distinct functional roles~\cite{voita2019mhsa}, softmax assigns strictly positive weight to \emph{every} key, so even a head tuned to local structure must still aggregate a contribution from every token~\cite{vaswani2017attention,peters2019sparse}. Each patch update therefore becomes a global mixture in which a long tail of low-relevance background and non-class tokens contributes to the local features. This motivates our central question: \textit{can a sparse attention distribution surface the localized, fine-grained cues that dense softmax masks?}

We instantiate this sparse alternative with ($\alpha$)-entmax~\cite{blondel2019learning}, which applies a data-dependent threshold that maps low-relevance attention scores \textit{exactly to zero} while preserving the probability simplex. \textit{Because it adds no learnable parameters, leaves the architecture unchanged, and smoothly generalizes softmax---recovering it as ($\alpha \to$  1)---entmax isolates attention sparsity as a single, principled variable.} In this work we present a systematic, training-free study of replacing softmax with $\alpha$-entmax in the final self-attention layers of a \emph{frozen} CLIP visual encoder---optionally paired with self-correlation attention scores---and evaluate its effect on dense, open-vocabulary prediction tasks, pixel-level semantic segmentation and region-level fine-grained recognition. We ask \emph{where} and \emph{why} attention sparsification helps, and show that attention sparsity is a relevant---yet largely overlooked---factor in addressing the limitations of CLIP to segmentation and fine-grained prediction tasks. We summarize our contributions as follows:

\begin{itemize}
  \item We empirically evaluate a training-free sparsifier for dense CLIP attention, replacing softmax with $\alpha$-entmax in the final attention layer(s) of a frozen encoder---adding no parameters and leaving the architecture and pretrained weights untouched.

  \item We analyze the known degradation from global \texttt{qk} mixing in CLIP's late attention layers through the lens of attention \emph{sparsity}, and establish our central finding: \textbf{the gain from sparsification is proportional to how diffuse the underlying attention is}---that is, how much each patch spreads its attention probability mass over non-class tokens rather than concentrating it on its own class. The gain is large on diffuse distributions and diminishes on already-concentrated ones, holding across the standard \texttt{qkv} attention and the self-correlated \texttt{qqv}, \texttt{kkv}, and \texttt{vvv} distributions.

  \item We show that entmax sparsity suppresses irrelevant global--local token mixing across segmentation (VOC, Context, ADE20K) and region--text retrieval (FG-OVD) on both ViT-B/16 and ViT-L/14, with the benefit increasing at higher resolutions and on larger backbones, which raise the token count and thus the global context each patch must attend over. 
\end{itemize}

\section{Related Work}
\paragraph{Dense Prediction with CLIP} The self-attention of CLIP's final encoder layer spreads probability mass broadly across the image, diminishing the per-patch discrimination that dense tasks require~\cite{mask-clip}. A body of training-free work converges on a common solution---keep CLIP frozen and modify the last block so each patch attends to itself and to similar patches rather than to global context---and differ mainly in \emph{what} each correlates. MaskCLIP~\cite{mask-clip} removes the query--key interaction and routes each patch's own value through the remaining projections. CLIP-Surgery~\cite{clip-surgery} substitutes value--value attention and subtracts the redundant component shared across class queries. SCLIP~\cite{sclip} sums query--query and key--key similarities, and GEM~\cite{gem} generalizes this into an ensemble of self-correlated attention over queries, keys, and values. ClearCLIP~\cite{lan2024clearclip} retains query--query attention while discarding the final block's residual and feed-forward branches. NACLIP~\cite{naclip} biases key--key attention toward spatial neighbours with a Gaussian kernel. CLIPtrase~\cite{shao2024cliptrace} averages cosine similarities across self-correlations and then applies clustering and heuristic denoising. These methods all change the \emph{inputs} to attention---the score source (query--key vs.\ self--self) or the value path---while leaving the \emph{softmax normalizer} in place. This suppresses the noisy tail distribution only indirectly, through hand-designed correlations and spatial priors that redistribute dense softmax mass \textit{rather than zeroing it out}.

\paragraph{Region-Text Alignment in CLIP}
Region-alignment work pools patch tokens over a bounding box---usually via \texttt{RoIAlign} on the feature map---and \emph{trains} that readout to carry region semantics. \cite{regionclip} aligns mined pseudo region--caption pairs contrastively, \cite{glip} adds a DETR-style detection-and-grounding head, \cite{wu2024clipselfvisiontransformerdistills} self-distills each patch toward its own crop embedding, and \cite{jing2024fineclip} and \cite{xie2025fgclip} extend this with region-caption supervision at scale. However, \cite{qiu2025refiningclip} shows that projecting patches into text space collapses their pairwise correlation structure even as retrieval and zero-shot accuracy hold---region--language training can obscure the spatial structure dense prediction needs. This problem is concentrated in CLIP's final layers, which its global matching objective shapes most strongly toward image-level alignment at the expense of localized structure. The fine-grained signal is therefore suppressed rather than lost, motivating an inference-time sparse intervention at those layers.

\paragraph{Entmax in Language Vision Models.}
Replacing softmax with entmax normalization in Transformer based models has been explored in prior work, most notably in language models to promote sparsity and interpretability. \cite{peters2019sparse} originally proposed replacing the softmax normalization in the output layer of a Transformer-based sequence-to-sequence decoder with entmax for $\alpha=1.5$. This substitution produced sparser token distributions and attention maps, assigning exact-zero weight to irrelevant source tokens and words and thereby making the true alignments emerge more distinctly.
Building on this, \cite{martins2020sparsetextgeneration} also replaces the softmax over the GPT-2 vocabulary logits during both training and decoding (for $\alpha$ = 1.2 1.3 and 1.5 depending on the dataset) while optimizing the corresponding entmax loss instead of cross-entropy.
\cite{correia19adaptively} pushed the substitution deeper into the network by replacing softmax with entmax in every multi-head self- and cross-attention layer while learning $\alpha$ separately for each head to encourage head-wise diversity. Inspired by entmax, \cite{zhou2025multimax} leaves softmax unchanged in a ViT, but precedes it with a learnable piece-wise temperature gate that pushes small logits toward zero while allowing several large peaks to remain active. In this work, we build on these foundations by replacing softmax with entmax in the later self-attention layers of the large language vision model CLIP.

\section{Method}
\label{sec:method}

We study a single, training-free modification to a frozen vision--language transformer: in the visual self-attention of the final layers(s) we replace the row-wise softmax with the $\alpha$-entmax transform. Because entmax maps the low-magnitude tail of each attention row exactly to zero, it acts as an implicit \emph{denoiser} over attention. Our goal is analysis rather than a new state-of-the-art approach: we ask \emph{where} and \emph{why} attention sparsification helps dense, open-vocabulary prediction.

\subsection{From Dense to Sparse Attention}
Self-attention in the original Transformer \cite{vaswani2017attention}
maps a sequence of token embeddings
$X=[x_1,\dots,x_N]\in\mathbb{R}^{N\times d_\text{model}}$
to a new sequence
$Z=[z_1,\dots,z_N]$ through learned linear projections
 
\[
Q = XW^{Q},\qquad
K = XW^{K},\qquad
V = XW^{V},
\qquad
W^{Q},W^{K},W^{V}\in\mathbb{R}^{d_\text{model}\times d_k}.
\]
 
The core computation---\emph{scaled dot-product attention}---combines these
matrices as
 
\[
\textstyle
Z
= \operatorname{Attention}(Q,K,V)
= \underbrace{\operatorname{softmax}\!\Bigl(
              \tfrac{QK^{\!\top}}{\sqrt{d_k}}
            \Bigr)}_{\displaystyle P\in\mathbb{R}^{N\times N}}\,V.
\]
 
Here, the matrix product $QK^{\!\top}$ yields the raw attention scores, which measure the alignment between each query and key token. A row-wise softmax is then applied to these scores, converting them into a probability distribution over the keys for each query. For an individual query vector $q_i$ (row $i$ of $Q$) the corresponding row of $P$ is the probability simplex vector
 
\[
p_i
= \operatorname{softmax}\!\Bigl(
    \tfrac{q_iK^{\!\top}}{\sqrt{d_k}}
  \Bigr)
\in\Delta^{N-1},
\qquad
z_i = p_i V = \sum_{j=1}^{N} p_{ij}\,v_j.
\]

Softmax can be characterized as the solution of a
Shannon-entropy--regularised linear objective
\cite[Def.~1]{blondel2019learning},\,
\cite[Eq.~7]{peters2019sparse}:
 
\[
\operatorname{softmax}(z)
=\arg\max_{p\in\Delta^{N-1}}
\Bigl[p^{\!\top}z + H(p)\Bigr],
\qquad
H(p)
=-\sum_{j}p_j\log p_j .
\]
 
The linear term $p^{\!\top}z$ pulls all mass onto the highest-scoring logit, while the entropy term opposes this concentration. 
Due to this entropy maximization, each query attends---however weakly---to every key. 
 
This variational formulation generalizes by allowing the Shannon entropy to be replaced with alternative regularizers, inducing sparser or otherwise structured attention distributions. When replacing the Shannon entropy with the Tsallis
\(\alpha\)-entropy, the optimization yields the \(\alpha\)-entmax transform derived by \cite[Eq.~(10)]{blondel2019learning} and adapted by
\cite[Eq. 9]{peters2019sparse}:
\[
  \operatorname{entmax}_{\alpha}(z)
  \;=\;
  \arg\!\max_{p \in \Delta^{N-1}}
  \Bigl[
      p^{\!\top} z
      + H_{\alpha}(p)
  \Bigr],
  \quad
  H_{\alpha}(p)
  = \frac{1 - \sum_{j} p_j^{\alpha}}
         {\alpha (\alpha - 1)},
  \quad
  \alpha \in (1, 2],
\]
 
where $\alpha \to 1$ recovers softmax and $\alpha = 2$ gives the maximally sparse sparsemax \cite{martins2016softmaxsparsemaxsparsemodel}.

Solving the objective yields a thresholded closed form,
\[
  p_j = \operatorname{entmax}_{\alpha}(z)_j
      = \bigl[(\alpha-1)z_j - \tau\bigr]_+^{\frac{1}{\alpha-1}},
  \qquad [\,x\,]_+ \equiv \max(0,x),
\]
in which a single scalar threshold $\tau$ separates the active coordinates from those mapped exactly to zero, and is set so that the surviving entries sum to one \cite{peters2019sparse, blondel2019learning}. The scalar $\tau$ is the Lagrange multiplier enforcing the normalization constraint $\mathbf 1^{\!\top} p = 1$.

The mechanism is clearest in the sparsemax case $\alpha = 2$, where the transform reduces to $p_j = [\,z_j - \tau\,]_+$. Entmax operates by sorting the scores $z_{(1)} \ge z_{(2)} \ge \cdots \ge z_{(N)}$ and finding the largest index $k$ such that the $k$-th score remains active. Because the surviving probabilities must sum to one, this cutoff condition reduces algebraically to checking whether the $k$-th sorted logit exceeds the threshold implied by the top $k$ scores:
\[
  1 + k\,z_{(k)} > \sum_{i=1}^{k} z_{(i)} .
\]
Once the active-set size $k$ is determined, the threshold $\tau$ is the shifted mean of those top scores:
\[
  \tau = \frac{\bigl(\sum_{i=1}^{k} z_{(i)}\bigr) - 1}{k}.
\]
Applying it clips the long tail of the distribution: any score $z_j \le \tau$ is mapped exactly to zero. For intermediate $\alpha$ the threshold generally has no closed form and is found by bisection \cite{peters2019sparse}. The exception is $\alpha = 1 + 1/n$ for integer $n$---notably $\alpha = 1.5$---where the exponent $1/(\alpha-1)$ is an integer and an exact sort-based algorithm again applies. In every case the support is selected by a single threshold, so the sparsification mechanism is unchanged.

\paragraph{Dimensional Scaling vs. Sparse Projection.}
As described in \cite{vaswani2017attention}, the factor $1/\sqrt{d_k}$ is necessary, which, if left unscaled, the large-magnitude logits drive softmax toward a saturated, near one-hot distribution. Notably, this raises the question of whether $\alpha$-entmax induced sparsity might simply lead to this saturated regime. In contrast, entmax achieves sparsity through a principled, data-dependent threshold $\tau$ that acts as a selective noise gate, truncating the long tail of low-scoring tokens, which behaves very differently than large-magnitude logits \emph{saturating the maximum}.




\subsection{Attention Probability Distributions}
In standard attention, the probability distribution is derived from the query--key dot product, $qk$, determining which tokens each query aggregates. Beyond the native query--key scores, we consider \emph{self-correlation} scores, which have been shown to localize better for dense prediction \cite{clip-surgery,gem,sclip,lan2024clearclip,naclip,shao2024cliptrace}. For a single modified block we compute raw attention scores $A$ as a direct swap of the query--key product, reusing the block's own projections and the standard per-head
scale $1/\sqrt{d}$:
\begin{itemize}\itemsep2pt
  \item \texttt{qkv}: the native $A=\phi(QK^{\top}/\sqrt{d})$ (standard attention)
  \item \texttt{qqv}/\texttt{kkv}/\texttt{vvv}: a self-correlation
        $A=\phi(SS^{\top}/\sqrt{d})$ with $S\in\{Q,K,V\}$, aggregating $V$---i.e.\ the
        query--key product is replaced by a self-product of one projection, with the
        same scaling and architecture as the $qk$ baseline
  \item \texttt{qq+kk} and \texttt{qq+kk+vv}: the mean of the corresponding
        self-correlation matrices, $\tfrac1m\sum_{S}\phi(SS^{\top}/\sqrt{d})$,
        as in the underlying distribution in \cite{sclip} and \cite{gem}
\end{itemize}
where $\phi$ is the row normalizer (softmax at $\alpha{=}1$, $\alpha$-entmax otherwise) and $d$ is the per-head dimension. In every case the aggregated tokens pass through the original output projection, residual, and MLP. Following prior findings that the last layers are the most effected by the global objective, we sparsify only the last few layers, allowing the earlier layers to more closely preserve the broad contextual representations learned during pretraining. 

\section{Evaluation}
We evaluate on two complementary training-free, open-vocabulary tasks: dense semantic segmentation (pixel-level) and fine-grained region classification (region-level). 
In each case, the dense features are the final-layer patch tokens after the post-LayerNorm and the output projection into the joint image--text space. Positional embeddings are resampled to the input grid, so the full image is encoded at its native resolution in a single forward pass. We study two frozen backbones \textbf{CLIP ViT-B/16} and \textbf{CLIP ViT-L/14}.

We additionally evaluate a reference configuration,  which removes the final block's query--key mixing and retains only the value path. This approach was initially introduced as part of the MaskCLIP~\cite{mask-clip} model, and has since become a widely adopted approach for evaluation across methods \cite{zohra2026-clip-83b, xie2025fgclip}.

\subsection{Open-Vocabulary Semantic Segmentation}
We follow the \cite{gem}'s evaluation protocol. Each image is resized to a fixed short side (divisible by the patch size) and encoded in one pass. We compute the cosine similarity between every patch embedding and the text embeddings of the class names (prompt ``\texttt{a photo of a \{class\}.}''), producing a class score map that is bilinearly upsampled to the image resolution. We report mIoU on \textbf{Pascal VOC} \cite{pascalvoc} (21 classes, with a background class predicted by thresholding the per-pixel softmax confidence), \textbf{Pascal Context} \cite{pascalcontext} (59 classes, foreground-only), and \textbf{ADE20K} \cite{ade20k} (150 classes, foreground-only). For VOC we sweep the background threshold and report the best. 

\subsection{Fine-Grained Open-Vocabulary Detection (FG-OVD)}
To probe whether the same effect holds for region-level recognition we use the FG-OVD benchmark~\cite{bianchi2023devil-cdf}, which stresses fine-grained attribute discrimination. For each ground-truth box we extract the dense feature map, \texttt{RoIAlign}~it over the box to a single region vector, and rank the positive caption against ten hard negatives obtained by swapping a single attribute (color, material, pattern, etc.). A sample is correct when the positive caption is top-1. We report top-1 accuracy on the \emph{hard}, \emph{medium}, and \emph{easy} attribute-difficulty splits and their mean (mHME), plus the \emph{trivial} shuffled-negative split. Captions are encoded verbatim (no prompt template), and images use each backbone's native preprocessing and resolution.

\section{Results}

\subsection{Sparse Distributions for Segmentation}

Table~\ref{tab:combined_gains} applies a single entmax layer and reports the gain over softmax for every distribution on both backbones. For the standard \texttt{qkv} attention we observe consistent gains, but since this is a poor grouping operator---a patch query attends to whatever is globally salient rather than to its own region---entmax can only reduce the tail of an already-noisy distribution, so the score stays low even where the relative gain is large (ADE, B/16: 1.74$\rightarrow$2.16). The self-correlation distributions (\texttt{qqv}, \texttt{kkv}, \texttt{qq+kk}) instead group similar patches, which already yeilds stronger localization. On these, entmax improves consistently across datasets and backbones, and most where dense softmax performs worst (\texttt{qqv} on ADE: $5.69\!\to\!12.9$). 

The value correlation \texttt{vvv} is the only distribution entmax degrades on ViT-B/16 (VOC/Context/ADE all negative) yet gains on ViT-L/14. We find that the reason for this, is because \texttt{vvv} already concentrates probability mass on the relevant tokens very well as show in (Fig.~\ref{fig:teaser}). Thus, it lacks a low-relevance tail to prune and thresholding results in discarding useful mass, whereas on the more spread-out L/14 even \texttt{vvv} carries removable noise. 
\textit{Sparsification therefore helps in proportion to how much the baseline attention spreads off the target class, and offers no benefit---or degrades performance---where that mass is already concentrated.}

\begin{figure}[th]
    \centering
    \begin{subfigure}{0.37\linewidth}
        \centering
        \adjustbox{trim=0 0 0 0, clip, width=0.75\linewidth}{%
        \includegraphics[width=\linewidth]{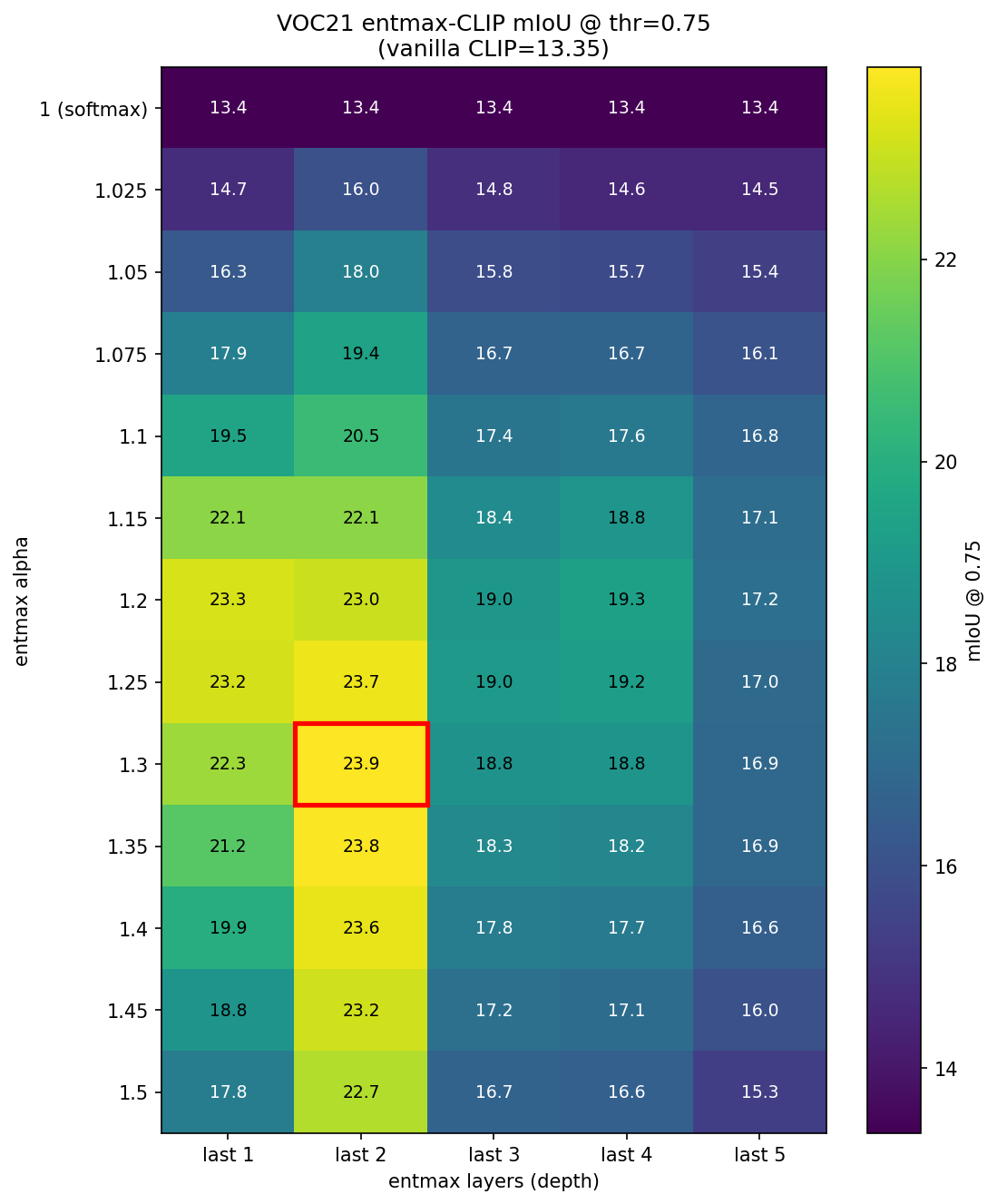}
        }
        \caption{\textbf{Sparsity vs.\ Depth on ViT-B/16 for \texttt{qkv} attention.} Sweeps $\alpha$ applied to the last $d$ transformer layers.}
        \label{fig:voc_alpha_depth}
    \end{subfigure}
    \begin{subfigure}{0.62\linewidth}
    \centering
    \adjustbox{trim=0 0 0 0, clip, width=0.75\linewidth}{%
    \includegraphics{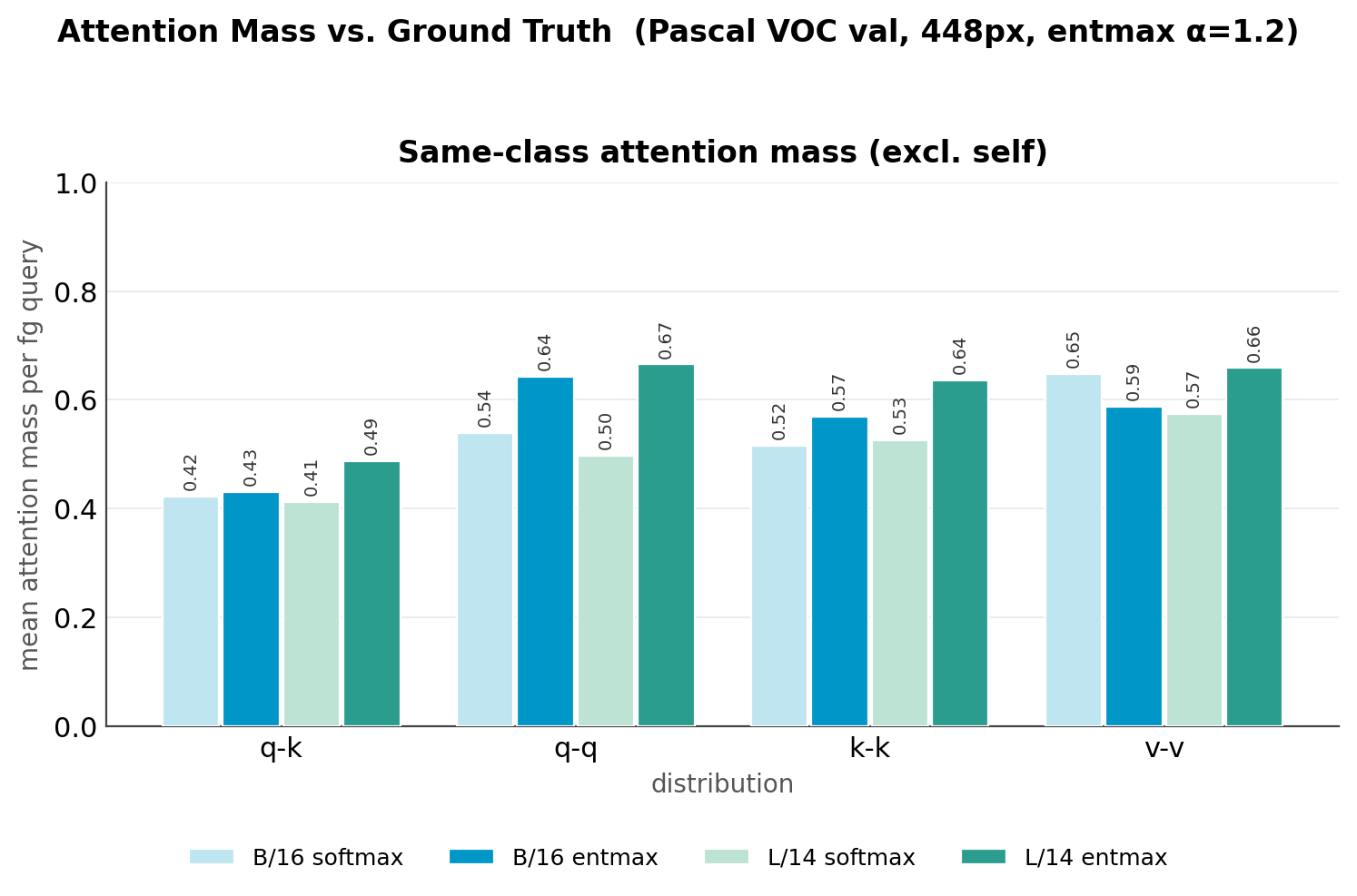}%
    }
    \caption{\textbf{Attention Mass in Softmax vs.\ Entmax Distributions.} Measures mean attention mass on \emph{same-class} patches, using the segmentation GT downsampled to the patch grid, excluding the self/diagonal mass.}
    \label{fig:same_class_mass}
    \end{subfigure}
\end{figure}

We further observe this zeroing of mass on relevant tokens in Figure \ref{fig:same_class_mass}, in which for every foreground patch we measure the mean attention mass it places on \emph{same-class neighbour} patches. Across model sizes and distributions, entmax increases this same-class mass, showing that it redistributes attention toward the relevant tokens rather than merely sparsifying. The single exception is \texttt{vvv} on ViT-B/16, where same-class mass drops---the same already-concentrated distribution that entmax degrades in Table~\ref{tab:combined_gains}, and for the same reason.

\begin{table*}[th]
\centering
\begin{small}
\scalebox{0.80}{
\begin{tabular}{l cc>{\columncolor{gray!15}}c cc>{\columncolor{gray!15}}c cc>{\columncolor{gray!15}}c}
\toprule
& \multicolumn{3}{c}{VOC} & \multicolumn{3}{c}{Context} & \multicolumn{3}{c}{ADE} \\
\cmidrule(lr){2-4}\cmidrule(lr){5-7}\cmidrule(lr){8-10}
Dist. & sm & ent & $\Delta$ & sm & ent & $\Delta$ & sm & ent & $\Delta$ \\
\midrule
\multicolumn{10}{l}{\textit{ViT-B/16}} \\
\midrule
\texttt{qkv}       & 13.35 & 23.25 & \textbf{+9.9}  &  8.11 & 10.28 & \textbf{+2.2} &  1.74 &  2.16 & \textbf{+0.4} \\
\texttt{qqv}      & 33.21 & 42.58 & \textbf{+9.4}  & 21.77 & 27.77 & \textbf{+6.0} &  5.69 & 12.90 & \textbf{+7.2} \\
\texttt{kkv}      & 21.33 & 30.08 & \textbf{+8.8}  & 18.86 & 21.04 & \textbf{+2.2} &  4.81 &  9.05 & \textbf{+4.2} \\
\texttt{vvv}      & 45.27 & 42.52 & -2.8           & 28.43 & 27.02 & -1.4          & 13.20 & 12.72 & -0.5 \\
\texttt{qq+kk}    & 27.68 & 38.07 & \textbf{+10.4} & 21.36 & 25.85 & \textbf{+4.5} &  5.28 & 11.82 & \textbf{+6.5} \\
\texttt{qq+kk+vv} & 40.25 & 42.60 & \textbf{+2.4}  & 26.27 & 27.58 & \textbf{+1.3} &  9.58 & 13.11 & \textbf{+3.5} \\
\midrule
\multicolumn{10}{l}{\textit{ViT-L/14}} \\
\midrule
\texttt{qkv}       &  6.48 & 12.64 & \textbf{+6.2}  &  3.91 &  7.31 & \textbf{+3.4} & 0.92 & 1.76 & \textbf{+0.8} \\
\texttt{qqv}      &  6.11 & 17.52 & \textbf{+11.4} &  4.45 & 11.47 & \textbf{+7.0} & 0.94 & 4.55 & \textbf{+3.6} \\
\texttt{kkv}      &  5.02 &  9.69 & \textbf{+4.7}  &  2.78 &  5.75 & \textbf{+3.0} & 0.49 & 1.85 & \textbf{+1.4} \\
\texttt{vvv}      & 24.55 & 30.92 & \textbf{+6.4}  & 16.33 & 20.23 & \textbf{+3.9} & 6.19 & 9.96 & \textbf{+3.8} \\
\texttt{qq+kk}    &  5.28 & 12.83 & \textbf{+7.6}  &  3.49 &  8.47 & \textbf{+5.0} & 0.65 & 3.00 & \textbf{+2.4} \\
\texttt{qq+kk+vv} &  8.74 & 19.83 & \textbf{+11.1} &  6.80 & 13.36 & \textbf{+6.6} & 1.47 & 5.41 & \textbf{+3.9} \\
\bottomrule
\end{tabular}}
\end{small}
\vspace{0.6em}
\caption{\textbf{The denoising gain generalizes across datasets and model scales with attention distributions} with a single entmax layer $L_1$ for softmax ($\alpha{=}1.0$, sm) vs.\ entmax ($\alpha{=}1.2$, ent) at resolution 448. The self-correlations gain monotonically on both CLIP backbones, with entmax multiplying the score several-fold on datasets where vanilla dense collapses (e.g.\ ADE). The value correlation (\texttt{vvv}) gains only on the highly diffuse ViT-L/14.}
\label{tab:combined_gains}
\end{table*}

\subsection{Sparse Distributions for Region--Text Alignment}
Table~\ref{tab:fgovd} shows the comparison on FG-OVD region--text retrieval, showing the effect is not specific to per-pixel segmentation, but on tasks that evaluate patch-level features. We observe that the \texttt{qkv} distribution is not effected as much (B/16: $4.44 \to 4.22$ mHME)---the same ceiling we see in segmentation, and for the same reason: denoising a global-salience distribution interferes with the locality region retrieval needs. The self-correlated attention distributions gain sharply on both backbones (\texttt{qqv} on ViT-B/16: $9.96 \to 17.50$ and on ViT-L/14: $8.64 \to 16.13$), and the trivial-split numbers rise in step showing that the sharpened attention helps even the easiest discrimination rather than trading one regime for another. 

\begin{table*}[th]
\centering
\resizebox{0.95\linewidth}{!}{%
\setlength{\tabcolsep}{5pt}
\begin{tabular}{ll cccc>{\columncolor{gray!10}}c c cccc>{\columncolor{gray!10}}c}
\toprule
& & \multicolumn{5}{c}{\textbf{Softmax}} & & \multicolumn{5}{c}{\textbf{Entmax}} \\
\cmidrule(lr){3-7} \cmidrule(l){9-13}
\textbf{Backbone} & \textbf{Dist} & \textbf{h} & \textbf{m} & \textbf{e} & \textbf{triv.} & \textbf{mHME} & & \textbf{h} & \textbf{m} & \textbf{e} & \textbf{triv.} & \textbf{mHME} \\
\midrule
\multirow{6}{*}{\textbf{ViT-B/16}}
 & \texttt{qkv}       &  2.45 &  5.86 &  5.00 & 14.36 & \cellcolor{gray!25}\textbf{4.44}  & &  2.28 &  5.22 &  5.16 & 18.53 &  4.22 \\
 & \texttt{qqv}      &  5.87 & 13.01 & 11.01 & 31.26 &  9.96          & & 10.97 & 22.20 & 19.32 & 55.83 & \textbf{17.50} \\
 & \texttt{kkv}      &  5.92 & 13.41 & 11.86 & 31.93 & 10.40          & & 10.13 & 19.81 & 17.32 & 50.35 & \textbf{15.75} \\
 & \texttt{vvv}      & 11.17 & 22.88 & 21.25 & 60.85 & 18.43          & & 11.85 & 24.12 & 23.09 & 60.90 & \cellcolor{gray!25}\textbf{19.69} \\
\addlinespace[2pt]
 & \texttt{qq+kk}    &  5.81 & 13.01 & 10.70 & 31.59 &  9.84          & & 10.52 & 21.23 & 18.86 & 54.19 & \textbf{16.87} \\
 & \texttt{qq+kk+vv} &  7.76 & 16.78 & 14.47 & 45.08 & 13.00          & & 11.37 & 22.88 & 20.94 & 57.69 & \textbf{18.40} \\
\midrule
\multirow{6}{*}{\textbf{ViT-L/14}}
 & \texttt{qkv}       &  5.59 &  8.76 &  6.16 &  6.21 &  6.84          & &  6.38 & 10.75 &  7.08 &  8.94 & \textbf{8.07} \\
 & \texttt{qqv}      &  6.74 & 10.34 &  8.85 &  8.35 &  8.64          & & 11.51 & 19.24 & 17.63 & 24.57 & \textbf{16.13} \\
 & \texttt{kkv}      &  7.28 & 11.15 & 11.32 &  9.42 &  9.92          & & 11.03 & 18.63 & 17.09 & 21.21 & \textbf{15.58} \\
 & \texttt{vvv}      & 11.71 & 20.59 & 18.17 & 35.43 & 16.82          & & 14.89 & 28.81 & 24.02 & 46.60 & \cellcolor{gray!25}\textbf{22.57} \\
\addlinespace[2pt]
 & \texttt{qq+kk}    &  6.74 & 10.85 & 10.01 &  8.63 &  9.20          & & 11.34 & 19.20 & 18.24 & 22.57 & \textbf{16.26} \\
 & \texttt{qq+kk+vv} &  8.29 & 13.54 & 11.47 & 14.16 & 11.10          & & 13.12 & 22.24 & 19.78 & 30.92 & \textbf{18.38} \\
\bottomrule
\end{tabular}%
}
\vspace{0.6em}
\caption{\textbf{The denoising gains transfer to fine-grained region--text retrieval on FG-OVD.} We compare the gains over softmax for entmax $\alpha{=}1.2$ applied at the last layer. We report the hard/medium/easy splits, the trivial shuffle, and their mean \textbf{mHME} over hard/medium/easy at resolution 224. Entmax denoising gives large, consistent gains on \texttt{qqv}, \texttt{qq+kk}, \texttt{qq+kk+vv}: $+5$--$7$ mHME on both backbones. }
\label{tab:fgovd}

\end{table*}

\subsection{Sparse Distributions at Shallow Depths}
Figure~\ref{fig:voc_alpha_depth} sweeps sparsity strength against the number of final layers entmax is applied to, and the \textit{optimum is at shallow depth and modest $\alpha$}. Increasing $\alpha$ reduces the gains showing that too much sparsification begins zeroing out relevant attention mass. Mild sparsity on the last 1-2 layers therefore lands in the regime where some global features are zeroed out, but the local features remain above the data-dependent $\alpha$-entmax threshold---consistent with the global objective concentrating its bias in the final layers.

\subsection{Effectiveness of Data-Dependent Sparsity}
We evaluate different sparse attentions in Table~\ref{tab:l14_methods}, which compares entmax against two controls at the last layer: a \emph{random} mask of matched support size, and temperature sharpening. On the noisy native \texttt{qkv} attention, entmax improves over softmax while the matched-support random mask that drops the same number of entries does not, showing the benefit comes from \emph{which} entries are pruned rather than from pruning alone. Temperature sharpening helps, but underperforms entmax, revealing that a data-dependent threshold is more beneficial than sharpness alone.

\begin{table}[t]
\centering
\resizebox{0.55\linewidth}{!}{%
\begin{tabular}{l cccc}
\toprule
Method & VOC & Context & ADE & FG-OVD \\
\midrule
\multicolumn{5}{l}{\textit{ViT-B/16 \texttt{qkv} attention normalization:}} \\
 \;\; softmax ($\alpha{=}1.0$)              & 13.44 &  8.02 & 1.66 & \textbf{4.44} \\
 \;\; temperature ($\tau{=}0.5$)           & 19.41 &  9.70 & 1.51 & 4.14 \\
 \;\; random (matched, $\alpha{=}1.2$)     & 13.38 &  8.00 & 1.66 & 4.40 \\
 \;\; entmax ($\alpha{=}1.2$)              & \textbf{21.75} & \textbf{10.16} & \textbf{1.79} & 4.22 \\
\midrule
\multicolumn{5}{l}{\textit{ViT-L/14 \texttt{qkv} attention normalization:}} \\
 \;\; softmax ($\alpha{=}1.0$)              &  6.06 &  3.83 & 0.89 & 6.84 \\
 \;\; temperature ($\tau{=}0.5$)           &  8.89 &  5.67 & 1.26 & 7.62 \\
 \;\; random (matched, $\alpha{=}1.2$)     &  6.05 &  3.81 & 0.89 & 6.84 \\
 \;\; entmax ($\alpha{=}1.2$)              & \textbf{10.88} & \textbf{6.63} & \textbf{1.54} & \textbf{8.07} \\
\bottomrule
\end{tabular}
}
\vspace{0.6em}
\caption{\textbf{The gain comes from content-aware noise removal, not from sparsity alone.} A matched-support random mask drops as many entries as entmax but ties softmax on the noisy \texttt{qkv} attention while temperature sharpening helps but trails entmax---isolating entmax's data-dependent threshold as the source of the gain. FG-OVD show the mean \textbf{mHME} over hard/medium/easy splits.}

\label{tab:l14_methods}
\end{table}

\subsection{Architectural Denoising}
\label{subsec:block}
Table~\ref{tab:drop_ablation_seg} ablates the final Transformer block, optionally dropping the post-attention FFN, the residual, or both (as in the ClearCLIP approach~\cite{lan2024clearclip}). On \texttt{qkv}, dropping the FFN alone does not effect the score much, whereas dropping the residual increases it sharply---showing that the residual significantly impedes localization by adding back the global features.

\begin{table*}[t]
\centering

\begin{small}
\scalebox{0.75}{
\begin{tabular}{ll cc>{\columncolor{gray!10}}c cc>{\columncolor{gray!10}}c cc>{\columncolor{gray!10}}c}
\toprule
& & \multicolumn{3}{c}{VOC} & \multicolumn{3}{c}{Context} & \multicolumn{3}{c}{ADE} \\
\cmidrule(lr){3-5}\cmidrule(lr){6-8}\cmidrule(lr){9-11}
Backbone & Last-block mod. & sm & ent & $\Delta$ & sm & ent & $\Delta$ & sm & ent & $\Delta$ \\
\midrule
\multicolumn{11}{l}{\textit{\texttt{qkv} attention}} \\
\midrule
\multirow{4}{*}{ViT-B/16}
 & full block           & 13.35 & 23.25 & \textbf{+9.9} &  8.11 & 10.28 & \textbf{+2.2} & 1.74 & 2.16 & \textbf{+0.4} \\
 & drop FFN             & 13.79 & 23.46 & \textbf{+9.7} &  8.46 & 11.34 & \textbf{+2.9} & 1.45 & 2.23 & \textbf{+0.8} \\
 & drop residual        & 30.41 & 36.24 & \textbf{+5.8} & 16.45 & 16.50 & \textbf{+0.1}          & 5.90 & 6.19 & \textbf{+0.3} \\
 & drop both            & 31.30 & 35.71 & \textbf{+4.4} & 21.28 & 21.79 & \textbf{+0.5}          & 8.20 & 9.54 & \textbf{+1.3} \\
\midrule
\multirow{4}{*}{ViT-L/14}
 & full block           &  6.48 & 12.64 & \textbf{+6.2} &  3.91 &  7.31 & \textbf{+3.4} & 0.92 & 1.76 & \textbf{+0.8} \\
 & drop FFN             &  7.44 & 14.62 & \textbf{+7.2} &  4.38 &  8.36 & \textbf{+4.0} & 1.06 &  2.00 & \textbf{+0.9} \\
 & drop residual        &  8.33 & 19.99 & \textbf{+11.7}&  5.67 &  9.92 & \textbf{+4.2} & 2.31 &  4.06 & \textbf{+1.7} \\
 & drop both            & 33.98 & 37.36 & \textbf{+3.4} & 21.19 & 23.50 & \textbf{+2.3} & 8.79 & 11.19 & \textbf{+2.4} \\
\bottomrule
\end{tabular}}
\caption{\textbf{Transformer Block Ablation.} On the \emph{final} layer Transformer block we optionally drop the post-attention FFN (\texttt{noffn}), the $+x$ residual(\texttt{nores}), or both (ClearCLIP-style)\cite{lan2024clearclip}. We evaluate Softmax ($\alpha{=}1.0$) vs.\ $\alpha$-entmax ($\alpha{=}1.2$) at resolution 448. On the \texttt{qkv} distribution, removing the \emph{residual}---not the FFN---lifts the performance sharply. Dropping the FFN alone does not effect \texttt{qkv} attention.}
\label{tab:drop_ablation_seg}
\end{small}
\end{table*}

\subsection{Scaling with Patch Count}
The resolution sweeps in Tables~\ref{tab:resolution} (FG-OVD) and~\ref{tab:resolution_seg_l14} (segmentation) show that increasing resolution, which corresponds to increases the number of patches increasingly benefits from sparse distributions. Since entmax removes a low-relevance tail, its advantage widens as that tail grows. On the self-correlated attention distributions the softmax baseline stagnates or declines as resolution rises while entmax \textit{keeps improving} as indicated by the $\Delta$ growing monotonically (\texttt{qqv} on B/16: $+7.5 \to +9.6 \to +10.5$). The same widening holds for L/14 segmentation across all three datasets, including \texttt{vvv}. Finally, entmax \texttt{vvv} surpasses the MaskCLIP-style value-only reference at every resolution (FG-OVD B/16: 23.03 vs.\ 22.36 at 448; segmentation L/14 VOC: 30.92 vs.\ 14.84), showing the gain comes from \emph{denoising} the value path rather than merely isolating it.

\begin{table}[h]
\centering
\resizebox{0.65\textwidth}{!}{
\begin{tabular}{ll rr>{\columncolor{gray!15}}r rr>{\columncolor{gray!15}}r rr>{\columncolor{gray!15}}r}
\toprule
\multicolumn{2}{c}{FG-OVD}& \multicolumn{3}{c}{224} & \multicolumn{3}{c}{336} & \multicolumn{3}{c}{448} \\
\cmidrule(lr){3-5} \cmidrule(lr){6-8} \cmidrule(lr){9-11}
Backbone & Dist. & sm & ent & $\Delta$ & sm & ent & $\Delta$ & sm & ent & $\Delta$ \\
\midrule
\multirow{5}{*}{ViT-B/16}
 & \texttt{qkv}  &  4.44 &  4.22 & $-0.2$ &  4.84 &  4.20 & $-0.6$ &  5.20 &  4.26 & $-0.9$ \\
 & \texttt{qqv} &  9.96 & 17.50 & \textbf{+7.5} &  9.87 & 19.50 & \textbf{+9.6} &  9.71 & 20.17 & \textbf{+10.5} \\
 & \texttt{kkv} & 10.40 & 15.75 & \textbf{+5.3} & 10.23 & 16.65 & \textbf{+6.4} &  9.85 & 17.34 & \textbf{+7.5} \\
 & \texttt{vvv} & 18.43 & \textbf{19.69 }& \textbf{+1.3} & 20.07 & \textbf{22.00} & \textbf{+1.9} & 20.94 & \textbf{23.03} & \textbf{+2.1} \\
 \cmidrule(lr){2-11}
 & \texttt{v-only} & \multicolumn{3}{c}{19.02} & \multicolumn{3}{c}{21.37} & \multicolumn{3}{c}{22.36} \\
\midrule
\multirow{5}{*}{ViT-L/14}
 & \texttt{qkv}  &  6.84 &  8.07 & \textbf{+1.2} &  6.54 &  8.02 & \textbf{+1.5} &  6.26 &  8.44 & \textbf{+2.2} \\
 & \texttt{qqv} &  8.64 & 16.13 & \textbf{+7.5} &  8.37 & 16.06 & \textbf{+7.7} &  8.07 & 16.59 & \textbf{+8.5} \\
 & \texttt{kkv} &  9.92 & 15.58 & \textbf{+5.7} &  9.14 & 14.41 & \textbf{+5.3} &  9.01 & 14.97 & \textbf{+6.0} \\
 & \texttt{vvv} & 16.82 & \textbf{22.57} & \textbf{+5.8} & 16.54 & \textbf{24.10} & \textbf{+7.6} & 16.82 &\textbf{ 25.22} & \textbf{+8.4} \\
 \cmidrule(lr){2-11}
 & \texttt{v-only} & \multicolumn{3}{c}{21.15} & \multicolumn{3}{c}{21.60} & \multicolumn{3}{c}{22.60} \\
\bottomrule
\end{tabular}%
}
\vspace{0.6em}
\caption{\textbf{The denoising gain grows with input resolution on fine-grained region--text retrieval on FG-OVD.} We evaluate mHME softmax $\alpha{=}1.0$ vs.\ entmax $\alpha{=}1.2$ for the last layer, and provide MaskCLIP-style value-only reference. As resolution rises the patch count grows quadratically ($14^2\!\to\!21^2\!\to\!28^2$ for /16, $16^2\!\to\!24^2\!\to\!32^2$ for /14), making softmax attention more spread out. For the self-correlated attention distributions (\texttt{qqv}, \texttt{kkv}, \texttt{vvv}) the softmax baseline stagnates while entmax keeps improving, so the gap $\Delta$ widens monotonically.}
\label{tab:resolution}
\end{table}

\begin{table}[h]
\centering
\resizebox{0.75\linewidth}{!}{ 
\begin{tabular}{ll cc>{\columncolor{gray!15}}c cc>{\columncolor{gray!15}}c cc>{\columncolor{gray!15}}c}
\toprule
& & \multicolumn{3}{c}{224} & \multicolumn{3}{c}{336} & \multicolumn{3}{c}{448} \\
\cmidrule(lr){3-5} \cmidrule(lr){6-8} \cmidrule(lr){9-11}
Dataset & Dist. & sm & ent & $\Delta$ & sm & ent & $\Delta$ & sm & ent & $\Delta$ \\
\midrule
Pascal VOC
 & \texttt{qkv}  &  6.06 & 10.88 & \textbf{+4.8} &  6.27 & 11.71 & \textbf{+5.4}  &  6.48 & 12.64 & \textbf{+6.2}  \\
 & \texttt{qqv} &  5.89 & 15.26 & \textbf{+9.4} &  5.96 & 16.33 & \textbf{+10.4} &  6.11 & 17.52 & \textbf{+11.4} \\
 & \texttt{kkv} &  5.08 &  9.11 & \textbf{+4.0} &  4.96 &  9.23 & \textbf{+4.3}  &  5.02 &  9.69 & \textbf{+4.7}  \\
 & \texttt{vvv} & \underline{23.25} & \textbf{28.19} & \textbf{+4.9} & \underline{23.94} & \textbf{29.93} & \textbf{+6.0}  & \underline{24.55} &\textbf{ 30.92} & \textbf{+6.4}  \\
 \cmidrule(lr){2-11}
& \texttt{v-only} & \multicolumn{3}{c}{13.40} & \multicolumn{3}{c}{14.08} & \multicolumn{3}{c}{14.84} \\

\midrule
Pascal Context
 & \texttt{qkv}  &  3.83 &  6.63 & \textbf{+2.8} &  3.82 &  6.97 & \textbf{+3.1} &  3.91 &  7.31 & \textbf{+3.4} \\
 & \texttt{qqv} &  4.76 & 11.03 & \textbf{+6.3} &  4.43 & 11.05 & \textbf{+6.6} &  4.45 & 11.47 & \textbf{+7.0} \\
 & \texttt{kkv} &  3.22 &  6.01 & \textbf{+2.8} &  2.84 &  5.65 & \textbf{+2.8} &  2.78 &  5.75 & \textbf{+3.0} \\
 & \texttt{vvv} & \underline{16.24} & \textbf{19.27} & \textbf{+3.0} & \underline{16.24} & \textbf{19.91} & \textbf{+3.7} & \underline{16.33} & \textbf{20.23} & \textbf{+3.9} \\
  \cmidrule(lr){2-11}
& \texttt{v-only} & \multicolumn{3}{c}{10.89} & \multicolumn{3}{c}{10.92} & \multicolumn{3}{c}{11.16} \\\midrule
ADE20K
 & \texttt{qkv}  & 0.89 & 1.54 & \textbf{+0.7} & 0.89 & 1.63 & \textbf{+0.7} & 0.92 & 1.76 & \textbf{+0.8} \\
 & \texttt{qqv} & 0.99 & 4.13 & \textbf{+3.1} & 0.94 & 4.26 & \textbf{+3.3} & 0.94 & 4.55 & \textbf{+3.6} \\
 & \texttt{kkv} & 0.60 & 1.96 & \textbf{+1.4} & 0.52 & 1.80 & \textbf{+1.3} & 0.49 & 1.85 & \textbf{+1.4} \\
 & \texttt{vvv} & \underline{6.29} & \textbf{8.99} & \textbf{+2.7} & \underline{6.11} & \textbf{9.68} & \textbf{+3.6} & \underline{6.19} & \textbf{9.96} & \textbf{+3.8} \\
  \cmidrule(lr){2-11}
& \texttt{v-only} & \multicolumn{3}{c}{5.38} & \multicolumn{3}{c}{5.53} & \multicolumn{3}{c}{5.69} \\
\bottomrule
\end{tabular}
}
\vspace{0.6em}
\caption{\textbf{The denoising gain grows with input resolution on dense segmentation.} We evaluate CLIP ViT-L/14 with resolutions $\{224,336,448\}$using softmax $\alpha{=}1.0$ vs.\ entmax $\alpha{=}1.2$ at the last layer. All distributions
gain with entmax and the gain $\Delta$ grows with resolution---including \texttt{vvv} ($+4.9\!\to\!+6.4$ on VOC). Because L/14's attention is diffuse throughout, there is always noise for entmax to remove.}
\label{tab:resolution_seg_l14}
\end{table}

\section{Qualitative Analysis}
We visualize the underlying dynamics of the attention distribution to explain the effect of redistribution of attention mass with entmax.

\begin{figure*}[h]
\centering
\begin{subfigure}{\linewidth}
\centering
\includegraphics[width=0.55\linewidth]{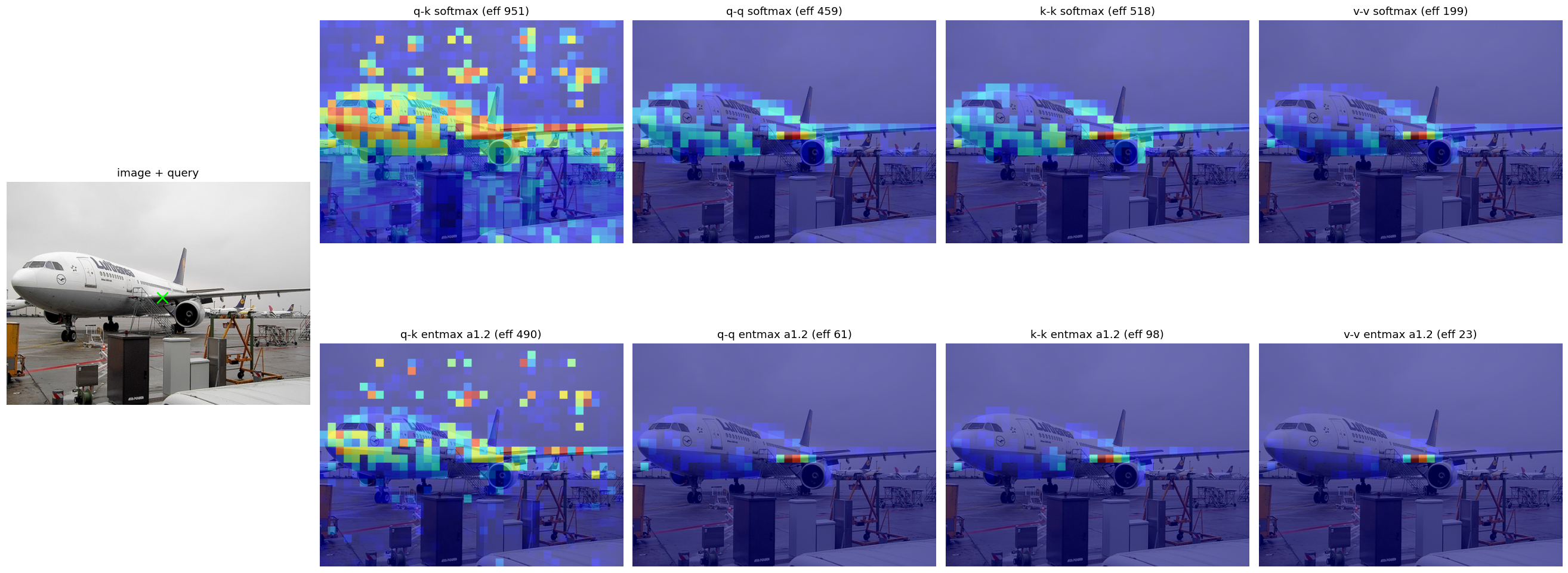}
\caption{Wing of an Airplane.}
\label{fig:diag_aeroplane}
\end{subfigure}
\vspace{0.6em}
\begin{subfigure}{\linewidth}
\centering
\includegraphics[width=0.55\linewidth]{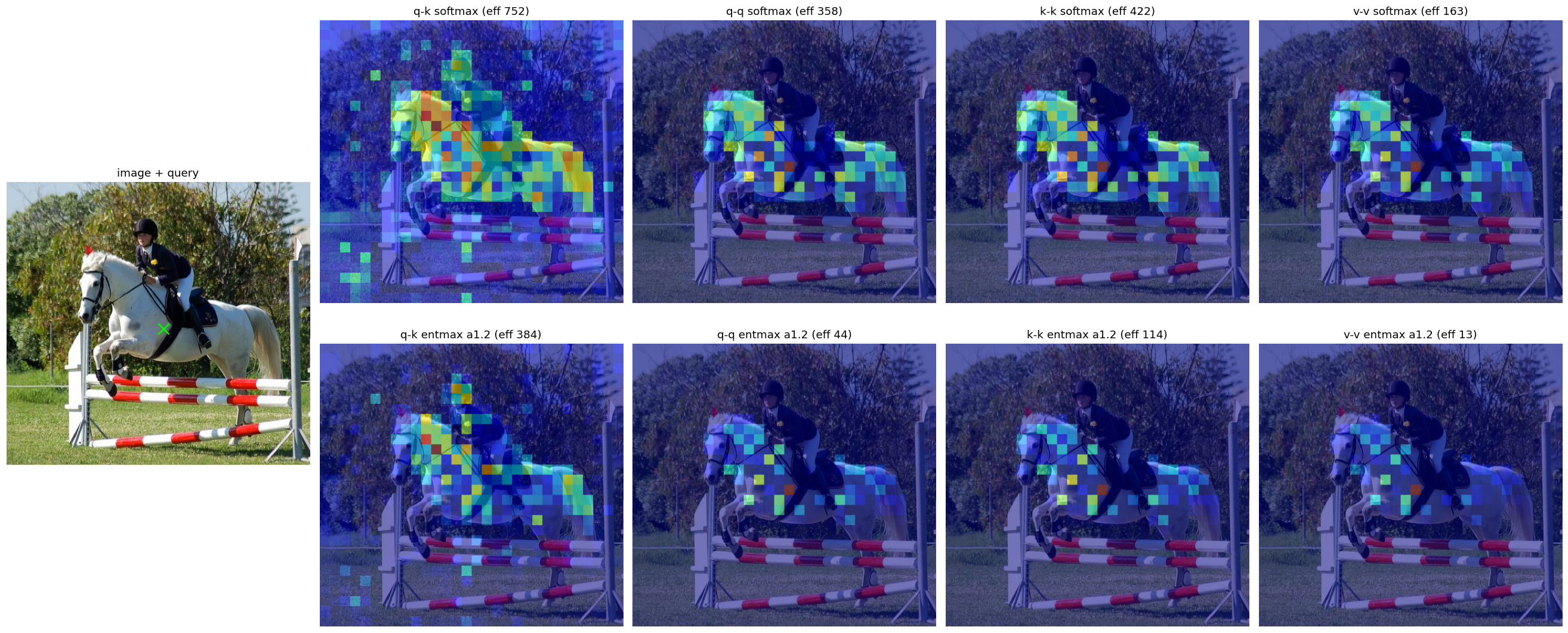}
\caption{Torso of a horse.}
\label{fig:diag_torso_horse}
\end{subfigure}







\caption{\textbf{Per-patch attention diagnostics.} CLIP ViT-B/16, last block. For a central query patch shown as (green~$\times$), we visualise the attention induced by each distribution---the standard \texttt{qk} and the self-correlated \texttt{qq}, \texttt{kk}, \texttt{vv}---under softmax (top row of each panel) and entmax ($\alpha{=}1.2$, bottom row). Titles report the effective number of attended keys. The \texttt{qk} attention is globally diffuse throughout the image, whereas the self-correlated attentions---especially \texttt{vv}---are far more locally coherent, concentrating on the object that contains the query. Entmax sharpens every distribution, narrowing the attention onto the coherent region and sharply reducing the effective key count.}
\label{fig:attn_diag}
\end{figure*}

\paragraph{Attention Distributions.} Figure~\ref{fig:attn_diag} traces a single central query patch through each score source under both normalizers. The \texttt{qk} attention is evidently globally diffuse, even with softmax, mass spreads well beyond the object and onto background, consistent with the last layer being shaped by the global matching objective. The self correlated attention distributions, on the other hand ---\texttt{qq}, \texttt{kk}, and especially \texttt{vv} concentrate on the object that contains the query before any sparsification is applied, showing why self-correlation alone is useful for localization. Swapping softmax for entmax $\alpha$=1.2 sharpens \emph{every} distribution, sharply reducing the effective number of attended keys reported in each image title.

\begin{figure*}[h]
    \centering
    \adjustbox{trim=0 0 0 0, clip, width=\linewidth}{%
    \includegraphics[]{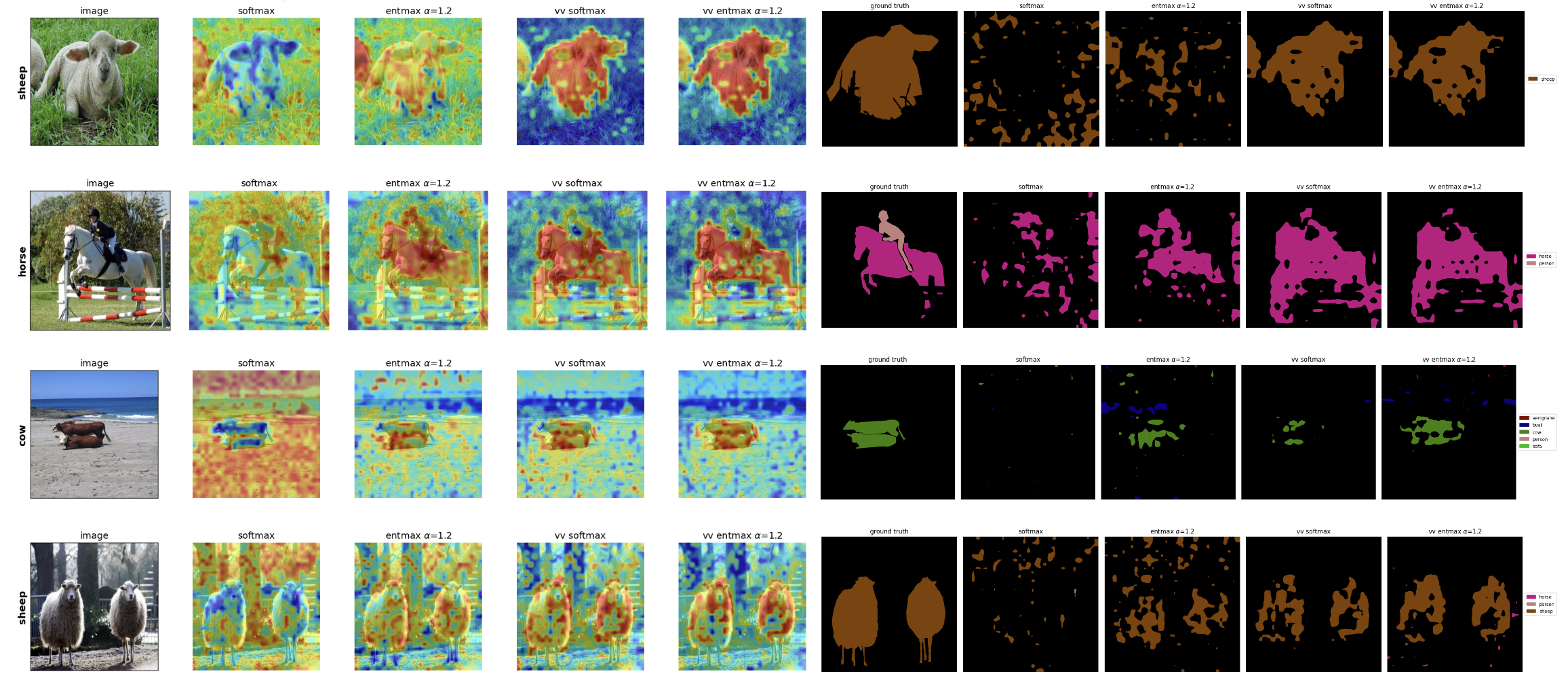}
    }

    \caption{\textbf{Qualitative Visualization of Segmentation Maps and Heatmaps.} The first two rows show visualizations for CLIP ViT-B/16 and the bottom two rows for ViT-L/14, at resolution $448$. The rows ordered by either the original/ground-truth, followed by softmax \texttt{qkv}, $\alpha$-entmax=1.2 \texttt{qkv}, softmax \texttt{vvv}, and $\alpha$-entmax=1.2 \texttt{vvv}.}
    \label{fig:segmaps_and_heatmaps}
\end{figure*}

\paragraph{Segmentation Maps and Similarity Heatmaps.} Figure~\ref{fig:segmaps_and_heatmaps} evaluates some examples of the dense prediction task. Assigning each patch its arg-max class (background at threshold of 0.85) and upsampling to the input resolution, the softmax maps spread class mass into surrounding background, reflecting the diffuse attention above. Entmax produces visibly tighter, more contiguous regions around objects. The improvement is qualitative evidence that the noise entmax removes is exactly the off-object attention mass that corrupts per-patch class assignments.

We also show the cosine similarity heatmaps between each patch embedding and a single class embedding. For a given class we take the cosine similarity between every patch embedding and that class embedding, reshape it to the patch grid, upsample to the image resolution, and min--max normalize. Under softmax the similarity is spatially spread out across the image. Entmax concentrates the similarity on the relevant region significantly better such that the peak aligns with the object rather than its surroundings, as in the case of the sheep, cow, and horse, which are primarily red vs. blue in their softmax counterparts. Notably, because these per-class fields are precisely what the segmentation maps threshold, the sharper localization here is the direct cause of the denoised features.

\section{Conclusion}We studied a single, training-free change to a frozen CLIP encoder---replacing the softmax normalizer in the final attention layers with $\alpha$-entmax to investigate how attention sparsification helps dense, open-vocabulary prediction. Because entmax adds no parameters and leaves the architecture and pretrained weights untouched, it isolates attention density as a single variable, separable from the score source the attention is computed over. That separation is our main finding. Across dense segmentation and fine-grained region--text retrieval, on both ViT-B/16 and ViT-L/14, sparsity consistently sharpens the self-correlated attention distributions by redistributing the low-relevance tail of locally irrelevant tokens that dense softmax spreads across every patch. We also show that this gain is proportional to how spread out the underlying distribution is---widening with higer resolution and with larger backbone as the noisy tail grows. 

\clearpage


\bibliographystyle{splncs04}
\bibliography{main}

\clearpage

\appendix

\section{Extended Quantitative Analysis}
 
\subsection{Resolution Scaling for Dense Segmentation on ViT-B/16}
 
Table~\ref{tab:resolution_seg} extends the resolution sweep to dense segmentation on ViT-B/16. The query- and key-driven localization behave as expected: \texttt{qqv}, \texttt{kkv}, and \texttt{qkv} on VOC all show a positive entmax gain $\Delta$ that grows or holds as the patch grid densifies, mirroring the FG-OVD trend. The effect is strongest on VOC, where every query/key localizer gains $+8$--$10$ mIoU at resolution $448$.
 
The value distribution \texttt{vvv} is the exception: it decreases slightly across all three segmentation benchmarks, and the deficit widens with resolution ($-2.0 \to -2.8$ on VOC). This mirrors the clean-versus-noisy signature from the main results --- on B/16 the value path is already locally coherent, so there is no diffuse tail for entmax to prune; sparsifying it instead risks cutting away relevant tokens.

\begin{table*}[th]
\centering
\resizebox{0.75\linewidth}{!}{ 

\begin{tabular}{@{} ll cc>{\columncolor{gray!15}}c cc>{\columncolor{gray!15}}c cc>{\columncolor{gray!15}}c @{}}
\toprule
& & \multicolumn{3}{c}{224} & \multicolumn{3}{c}{336} & \multicolumn{3}{c}{448} \\
\cmidrule(lr){3-5} \cmidrule(lr){6-8} \cmidrule(lr){9-11}
Dataset & Affinity & sm & ent & $\Delta$ & sm & ent & $\Delta$ & sm & ent & $\Delta$ \\
\midrule
\multirow{5}{*}{Pascal VOC}
 & \texttt{qqv} & 35.11 & 41.79 & \textbf{+6.7} & 34.36 & 43.09 & \textbf{+8.7} & 33.21 & \underline{42.58} & \textbf{+9.4} \\
 & \texttt{kkv} & 26.08 & 32.97 & \textbf{+6.9} & 23.53 & 31.54 & \textbf{+8.0} & 21.33 & 30.08 & \textbf{+8.8} \\
 & \texttt{qkv}  & 13.44 & 21.75 & \textbf{+8.3} & 13.31 & 22.74 & \textbf{+9.4} & 13.35 & 23.25 & \textbf{+9.9} \\
 & \texttt{vvv} & \textbf{44.01} & \underline{42.03} & $-2.0$ & \textbf{45.55} & \underline{43.17} & $-2.4$ & \textbf{45.27} & 42.52 & $-2.8$ \\
\cmidrule(lr){2-11}
 & \texttt{v-only} & \multicolumn{3}{c}{35.61} & \multicolumn{3}{c}{36.09} & \multicolumn{3}{c}{35.53} \\
\midrule
\multirow{5}{*}{Pascal Context}
 & \texttt{qkv}  &  8.02 & 10.16 & \textbf{+2.1} &  8.08 & 10.35 & \textbf{+2.3} &  8.11 & 10.28 & \textbf{+2.2} \\
 & \texttt{qqv} & 21.69 & \underline{26.66} & \textbf{+5.0} & 22.07 & \underline{27.88} & \textbf{+5.8} & 21.77 & \underline{27.77} & \textbf{+6.0} \\
 & \texttt{kkv} & 20.61 & 22.51 & \textbf{+1.9} & 19.99 & 22.10 & \textbf{+2.1} & 18.86 & 21.04 & \textbf{+2.2} \\
 & \texttt{vvv} & \textbf{26.81} & 26.17 & $-0.6$ & \textbf{28.25} & 27.19 & $-1.1$ & \textbf{28.43} & 27.02 & $-1.4$ \\
 \cmidrule(lr){2-11}
 & \texttt{v-only} & \multicolumn{3}{c}{23.32} & \multicolumn{3}{c}{23.80} & \multicolumn{3}{c}{23.36} \\

\midrule
\multirow{5}{*}{ADE20K}
 & \texttt{qkv}  &  1.66 &  1.79 & \textbf{+0.1} &  1.76 &  2.12 & \textbf{+0.4} &  1.74 &  2.16 & \textbf{+0.4} \\
 & \texttt{qqv} &  6.52 & \underline{12.19} & \textbf{+5.7} &  6.24 & \underline{12.97} & \textbf{+6.7} &  5.69 & \underline{12.90} & \textbf{+7.2} \\
 & \texttt{kkv} &  5.67 &  9.56 & \textbf{+3.9} &  5.32 &  9.59 & \textbf{+4.3} &  4.81 &  9.05 & \textbf{+4.2} \\
 & \texttt{vvv} & \textbf{12.43} & 12.12 & $-0.3$ & \textbf{13.32 }& 12.84 & $-0.5$ & \textbf{13.20} & 12.72 & $-0.5$ \\
 \cmidrule(lr){2-11}
 & \texttt{v-only} & \multicolumn{3}{c}{10.68} & \multicolumn{3}{c}{11.05} & \multicolumn{3}{c}{10.85} \\

\bottomrule
\end{tabular}
}
\caption{\textbf{Resolution scaling transfers to dense segmentation.} Evaluates ViT-B/16 for different resolutions with short-side $\{224,336,448\}$ using softmax $\alpha{=}1.0$ vs.\ entmax $\alpha{=}1.2$ at the last layer and provide MaskCLIP-style value-only for reference. For the query and key-based  distribution (\texttt{qqv}, \texttt{kkv}, and \texttt{qkv} on VOC) the entmax denoising gain $\Delta$ is positive and grows---or holds---as the patch grid gets denser. The effect is strongest on VOC.}
\label{tab:resolution_seg}
\end{table*}

\subsection{Ablating the FFN and Residual Pathway on FG-OVD}
 
Table~\ref{tab:drop_ablation_fgovd} isolates the contribution of the feed-forward network (FFN) and residual connection to final-layer \texttt{qkv} attention on fine-grained retrieval (FG-OVD). Removing the residual connection improves performance across both backbones, and entmax further improves over softmax on top of that gain. Dropping the FFN alone, by contrast, changes performance only marginally. Entmax consistently outperforms softmax on the larger ViT-L/14 model in particular. Peak accuracy --- 16.05 mHME on ViT-B/16 and 22.65 mHME on ViT-L/14 --- is reached by dropping the residual pathway and sparsifying the attention distribution together.

\begin{table*}[t]
\centering
\begin{small}
\scalebox{0.92}{
\begin{tabular}{ll ccc}
\toprule
Backbone & Last-block mod. & sm & ent & $\Delta$ \\
\midrule
\multicolumn{5}{l}{\textit{\texttt{qkv} attention}} \\
\midrule
\multirow{4}{*}{ViT-B/16}
 & full block    &  4.44 &  4.22 & -0.2 \\
 & drop FFN      &  5.28 &  4.85 & -0.4 \\
 & drop residual & 14.46 & 14.64 & +0.2 \\
 & drop both     & 15.59 & \textbf{16.05} & +0.5 \\
\midrule
\multirow{4}{*}{ViT-L/14}
 & full block    &  6.84 &  8.07 & +1.2 \\
 & drop FFN      &  6.29 &  7.45 & +1.2 \\
 & drop residual & 21.56 & \textbf{22.65} & +1.1 \\
 & drop both     & 20.07 & 22.51 & +2.4 \\
\bottomrule
\end{tabular}}
\end{small}
\caption{\textbf{Last Layer Ablation on FG-OVD.} Evaluates mHME for res $224$. On \texttt{qkv}, dropping the \emph{residual} increases the performance by a wide margin, while dropping the FFN alone is not as effective.}
\label{tab:drop_ablation_fgovd}
\end{table*}

\subsection{Sparsity--Depth on ViT-L/14}
 
We further evaluate the impact of $\alpha$-entmax across different depths and sparsity levels on ViT-L/14. As shown in Figure~\ref{fig:voc_alpha_depth}, the trends mirror those observed in the base architecture. For the \texttt{qkv} distribution, applying moderate sparsity ($\alpha \approx 1.3$) at the last two layers maximizes the performance gain. The more localized \texttt{vvv} distribution instead needs only mild sparsification ($\alpha{=}1.15$) at the final layer to reach peak accuracy. Notably, sparsifying any of the final three layers consistently improves performance for both attention variants.

\begin{figure}[h]
    \centering
    \begin{subfigure}{0.47\linewidth}
        \centering
        \includegraphics[width=\linewidth]{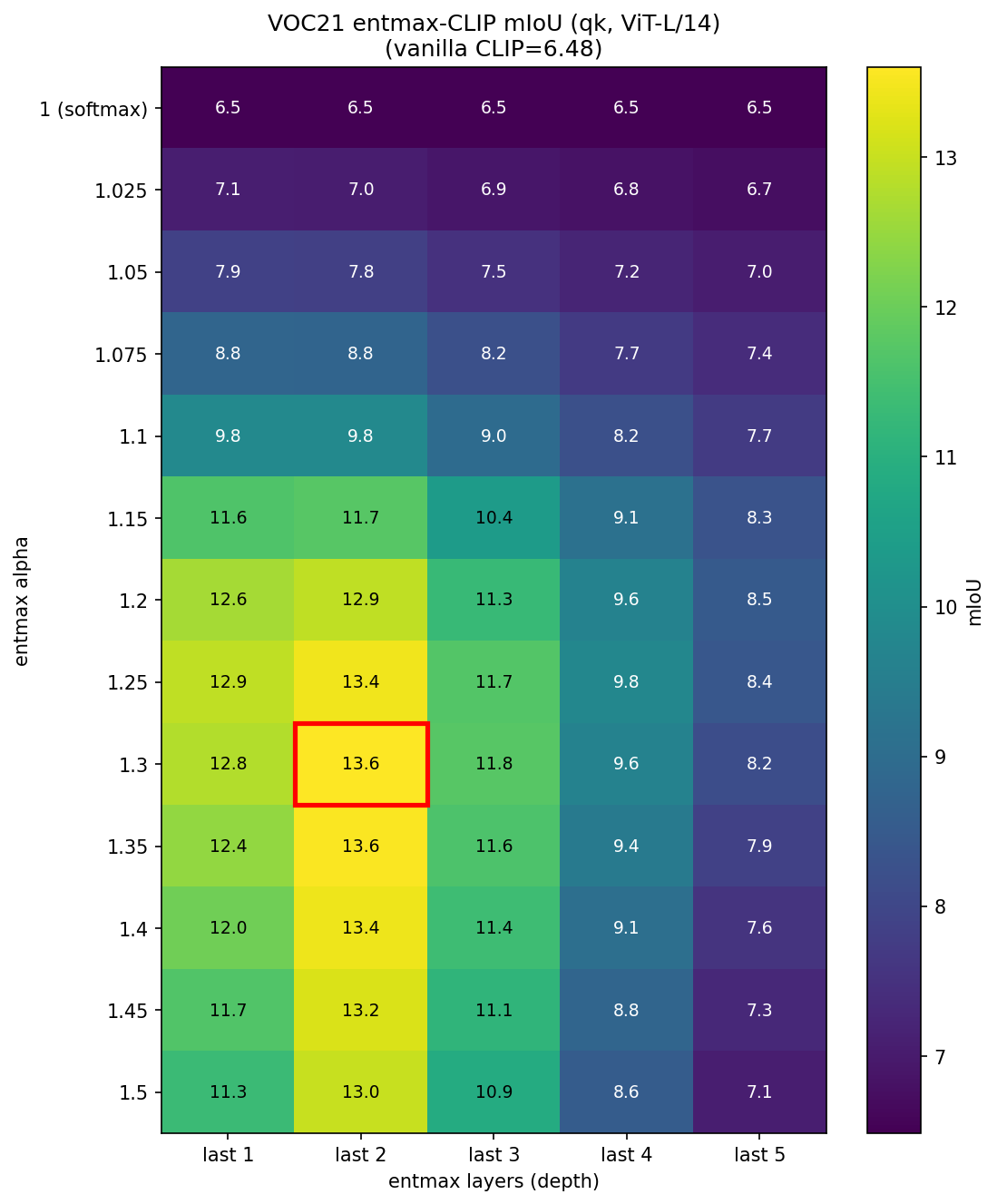} 
        \caption{ViT-L/14 \texttt{qkv} Attention}
        \label{fig:sub_right}
    \end{subfigure}
    \begin{subfigure}{0.47\linewidth}
        \centering
        \includegraphics[width=\linewidth]{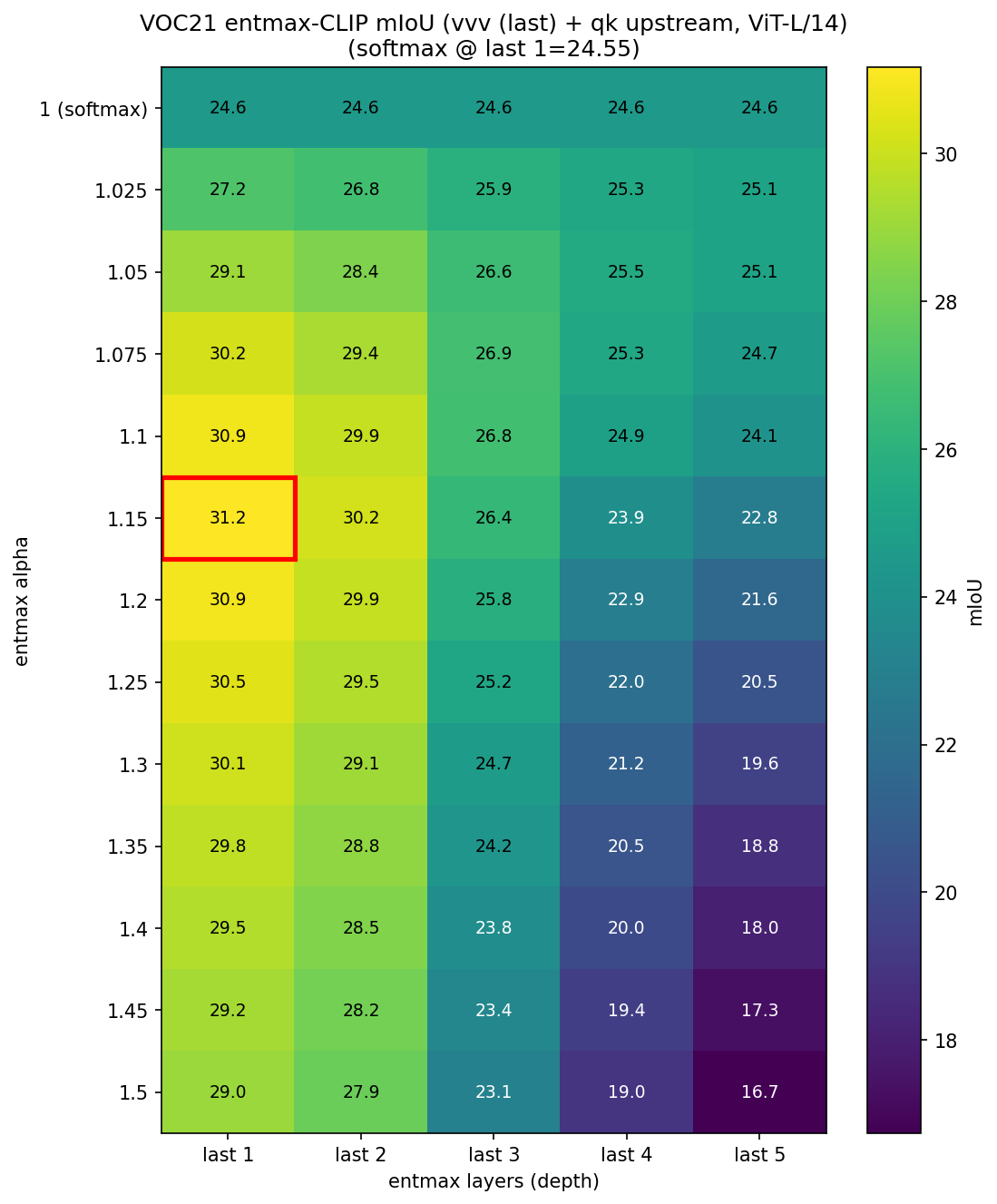} 
        \caption{ViT-L/14 \texttt{vvv} Attention}
        \label{fig:sub_right}
    \end{subfigure}
    \caption{\textbf{Entmax sparsity vs.\ depth on Pascal VOC on ViT-L/14} We evaluate the effectiveness of a sparse \texttt{qkv} attention distribution, across different depths and entropy regularization. Each cell reports dense-segmentation mIoU when entmax (column: sparsity $\alpha$) is applied to the last $d$ transformer blocks (row).}
    \label{fig:voc_alpha_depth}
\end{figure}

\section{Datasets and Benchmarks}
 
Table~\ref{tab:datasets_stats} summarizes key statistics of the datasets used for benchmarking.

\begin{table}[h]
\centering
\small
\scalebox{0.85}{
\begin{tabular}{@{}lcc@{}}
\toprule
Dataset & Classes / Queries & Images / Captions \\ 
\midrule
\multicolumn{3}{@{}l}{\textit{Task: Semantic Segmentation}} \\
Pascal VOC      & 21 (incl.\ bg) & 1.4k \\
Pascal Context  & 59 (fg-only)   & 5.0k \\
ADE20K          & 150 (fg-only)  & 2.0k \\
\midrule
\multicolumn{3}{@{}l}{\textit{Task: Region--Text Retrieval}} \\
FG-OVD (Hard)   & 2.3k boxes     & 1 pos.\ + 10 neg.\ each \\
\bottomrule
\end{tabular}
}
\caption{Statistics of the evaluation benchmarks.}
\label{tab:datasets_stats}
\end{table}

\section{Qualitative Attention Diagnostics}
Figures~\ref{fig:attn_diag} (this section) provide further per-patch attention diagnostics on ViT-B/16. Across diverse queries---a bicycle frame, a horse's front legs, a vase, a sofa, a boat, and a shirt collar---the pattern from the main paper recurs. The native \texttt{q-k} attention is globally diffuse and spreads onto background, the self-self distributions (especially \texttt{v-v}) already concentrate on the object containing the query, and entmax sharpens every distribution while sharply reducing the effective key count.

\begin{figure*}[t]
\centering

\vspace{0.6em}
\begin{subfigure}{\linewidth}
\centering
\includegraphics[width=\linewidth]{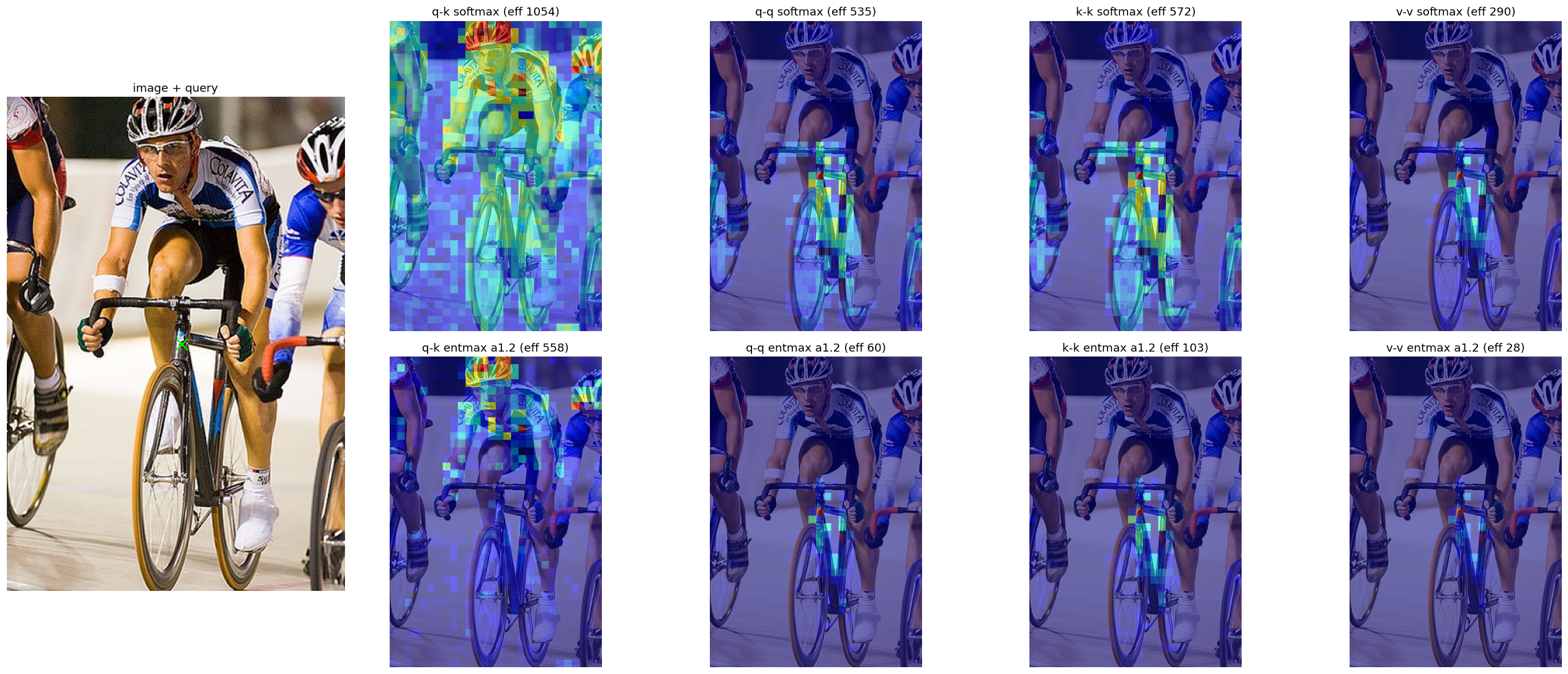}
\caption{Bicycle.}
\label{fig:diag_bicycle}
\end{subfigure}

\vspace{0.6em}
\begin{subfigure}{\linewidth}
\centering
\includegraphics[width=\linewidth]{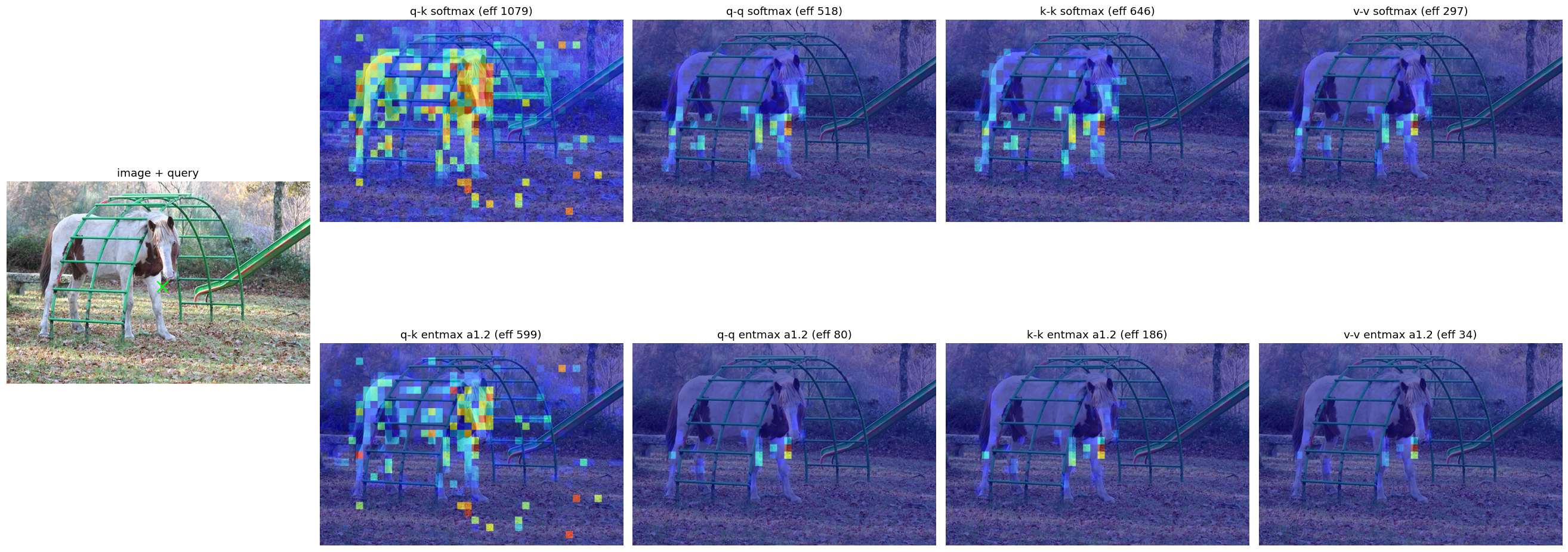}
\caption{Horse's front legs.}
\label{fig:diag_horse_legs}
\end{subfigure}

\vspace{0.6em}
\begin{subfigure}{\linewidth}
\centering
\includegraphics[width=\linewidth]{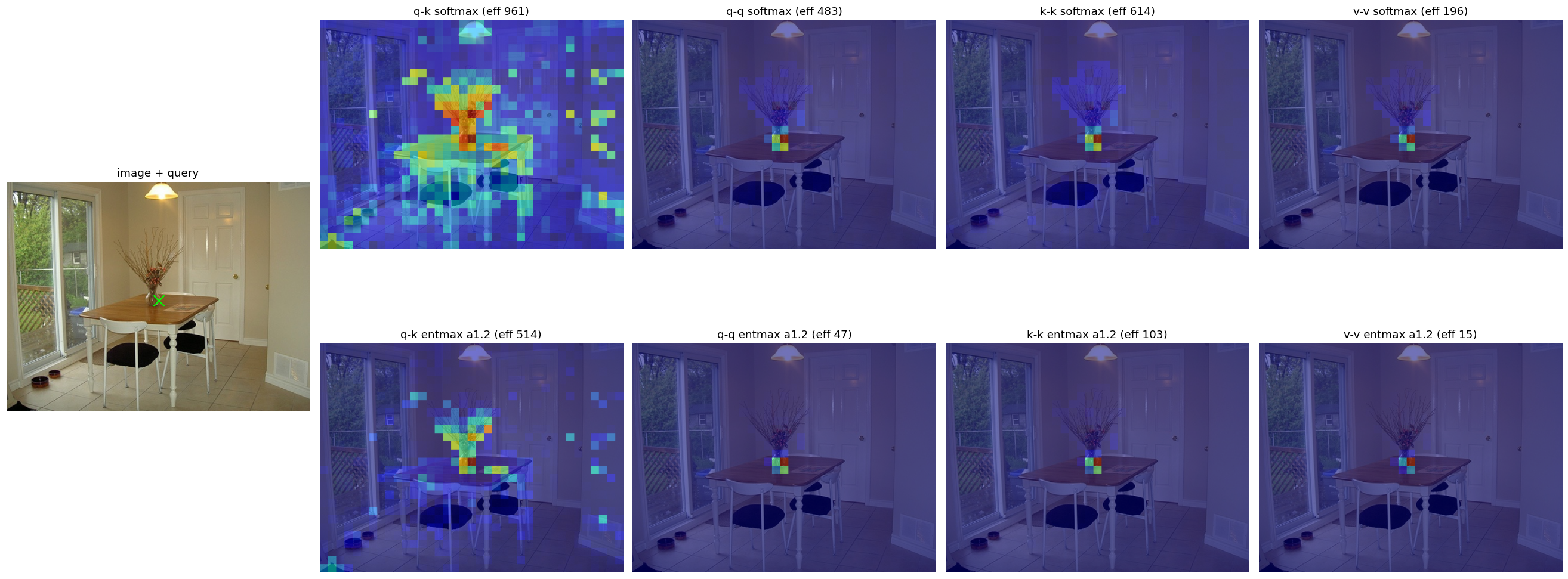}
\caption{Vase.}
\label{fig:vase.}
\end{subfigure}

\caption{\textbf{Additional per-patch attention diagnostics.} Show here for CLIP ViT-B/16, last block.}
\label{fig:attn_diag}
\end{figure*}

\begin{figure*}[t]
\centering

\vspace{0.6em}
\begin{subfigure}{\linewidth}
\centering
\includegraphics[width=\linewidth]{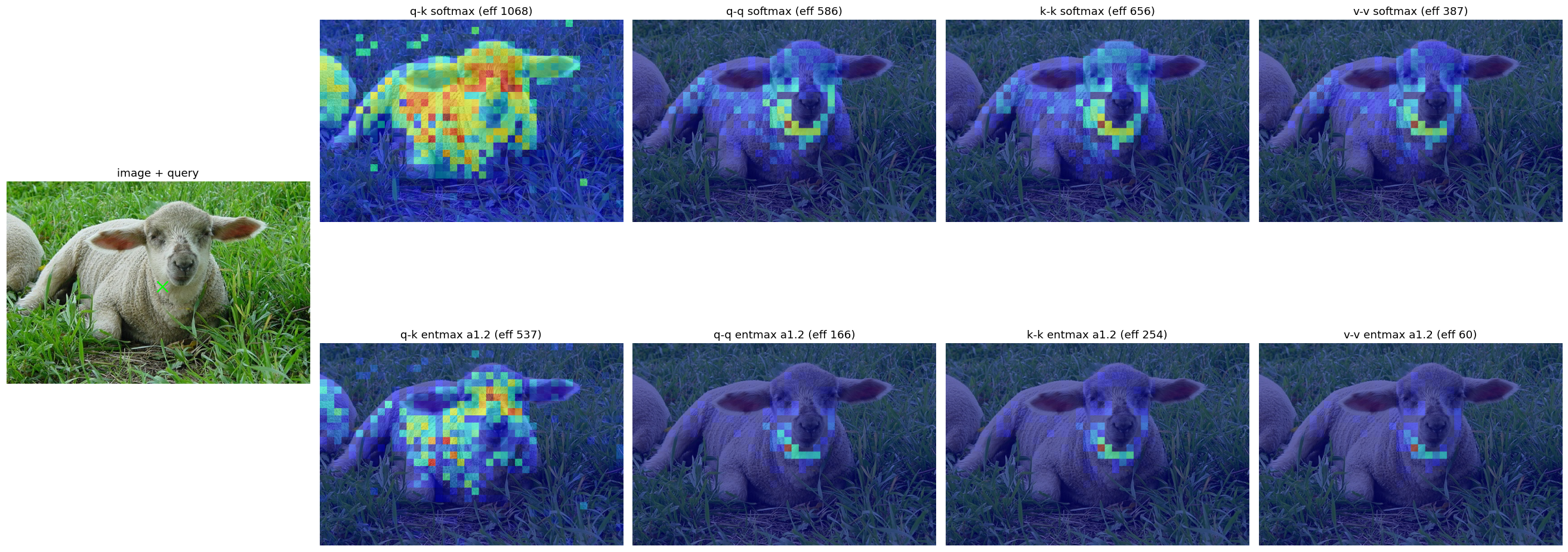}
\caption{Sheep's neck.}
\label{fig:diag_sheeps_neck}
\end{subfigure}



\vspace{0.6em}
\begin{subfigure}{\linewidth}
\centering
\includegraphics[width=\linewidth]{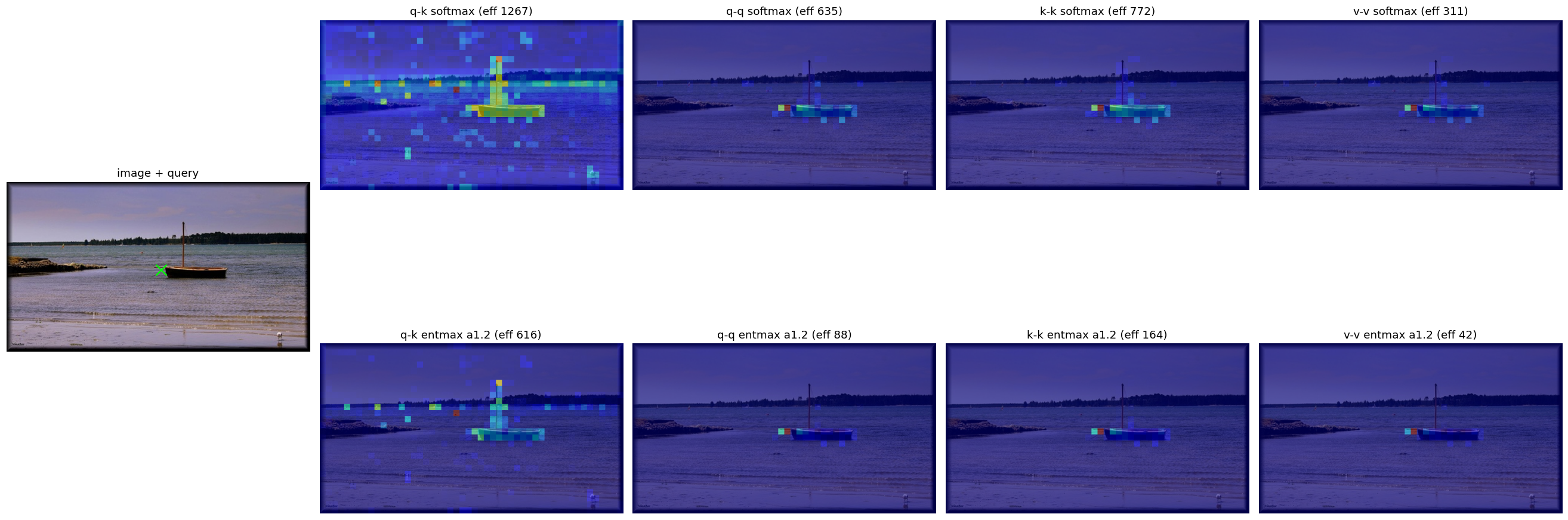}
\caption{Boat.}
\label{fig:boat}
\end{subfigure}

\vspace{0.6em}
\begin{subfigure}{\linewidth}
\centering
\includegraphics[width=\linewidth]{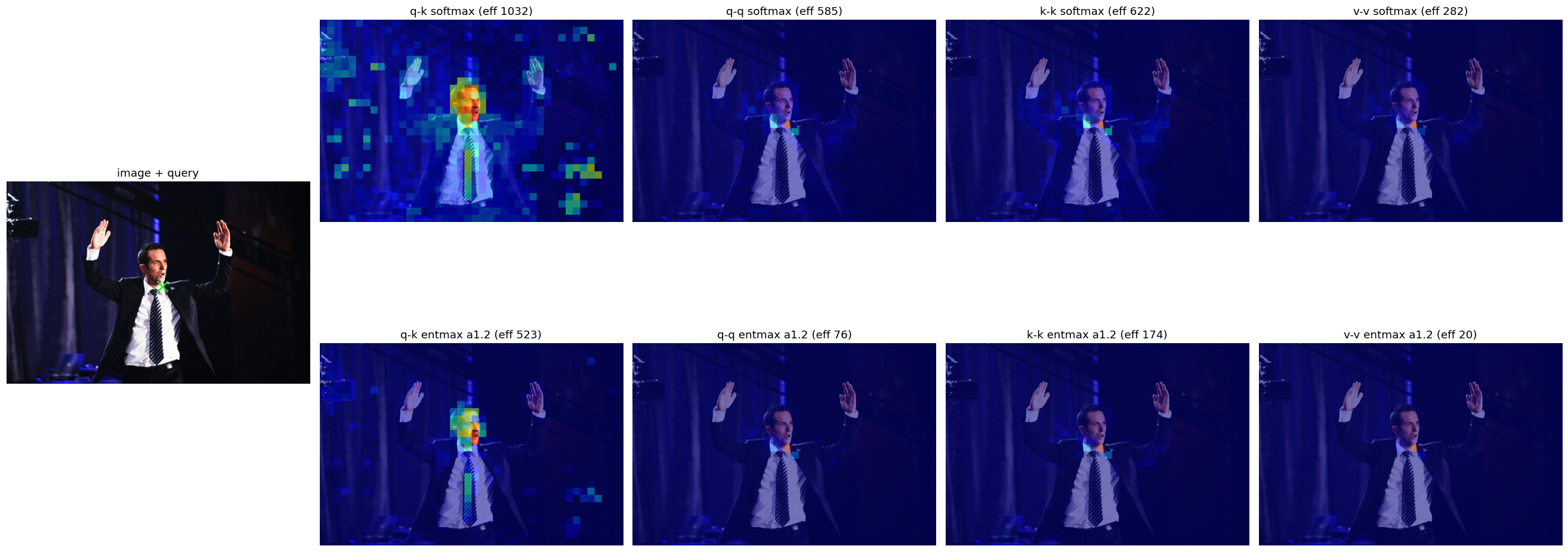}
\caption{Collar of shirt.}
\label{fig:collar_shirt}
\end{subfigure}

\caption{\textbf{Additional per-patch attention diagnostics.} Show here for CLIP ViT-B/16, last block.}
\label{fig:attn_diag}
\end{figure*}

\end{document}